\documentclass{article}

\PassOptionsToPackage{numbers, compress}{natbib}

\usepackage{amsmath}
\usepackage{amssymb}

\usepackage[final]{neurips_2022}

\usepackage[utf8]{inputenc} 
\usepackage[T1]{fontenc}    
\usepackage{hyperref}       
\usepackage{url}            
\usepackage{booktabs}       
\usepackage{amsfonts}       
\usepackage{nicefrac}       
\usepackage{microtype}      
\usepackage{xcolor}         
\usepackage{algorithmicx,algorithm}
\usepackage{multirow}
\usepackage[noend]{algpseudocode}
\usepackage{wrapfig}

\usepackage{graphicx}
\usepackage{svg}

\usepackage[toc,page,header]{appendix}
\usepackage{minitoc}

\title{MuSe-GNN: Learning Unified Gene Representation From Multimodal Biological Graph Data}

\author{%
  Tianyu Liu \\
  Yale University\\
  \texttt{tianyu.liu@yale.edu} \\
  \And
  Yuge Wang \\
  Yale University\\
  \texttt{yuge.wang@yale.edu} \\
  \And
  Rex Ying \\
  Yale University\\
  \texttt{rex.ying@yale.edu} \\
  \And
  Hongyu Zhao*\\
  Yale University\\
  \texttt{hongyu.zhao@yale.edu} \\
  *: Corresponding author
}

\begin{document}
\doparttoc 
\faketableofcontents 

\maketitle

\begin{abstract}
Discovering genes with similar functions across diverse biomedical contexts poses a significant challenge in gene representation learning due to data heterogeneity. In this study, we resolve this problem by introducing a novel model called Multimodal Similarity Learning Graph Neural Network, which combines Multimodal Machine Learning and Deep Graph Neural Networks to learn gene representations from single-cell sequencing and spatial transcriptomic data. Leveraging 82 training datasets from 10 tissues, three sequencing techniques, and three species, we create informative graph structures for model training and gene representations generation, while incorporating regularization with weighted similarity learning and contrastive learning to learn cross-data gene-gene relationships. This novel design ensures that we can offer gene representations containing functional similarity across different contexts in a joint space. Comprehensive benchmarking analysis shows our model's capacity to effectively capture gene function similarity across multiple modalities, outperforming state-of-the-art methods in gene representation learning by up to 97.5\%. Moreover, we employ bioinformatics tools in conjunction with gene representations to uncover pathway enrichment, regulation causal networks, and functions of disease-associated or dosage-sensitive genes. Therefore, our model efficiently produces unified gene representations for the analysis of gene functions, tissue functions, diseases, and species evolution. 
\end{abstract}
\section{Introduction}
Progress in biological technology has broadened the variety of biological data, facilitating the examination of intricate biological systems. A prime example of such technology is single-cell sequencing, which allows for the comprehensive characterization of genetic information within individual cells \cite{han2020construction, saliba2014single}. This technology provides access to the full range of a cell's transcriptomics, epigenomics and proteomics information, including gene expression (scRNA-seq) \cite{hwang2018single, zheng2017massively}, chromatin accessibility (scATAC-seq) \cite{cusanovich2015multiplex, chen2018rapid}, methylation \cite{luo2017single, mulqueen2018highly} and anti-bodies \cite{stoeckius2017simultaneous}. By sequencing cells from the same tissue at various time points, we can gain insight into patterns of cellular activity over time \cite{fischer2019inferring}. Moreover, spatial information of cellular activity represents an equally vital, additional dimension. Such data are defined as spatial data. All of these data are known as multi-omics data, and integrating multi-omics data for combined analysis poses a significant challenge.

\begin{figure}[ht]
    \centering
    \includegraphics[width=0.85\textwidth]{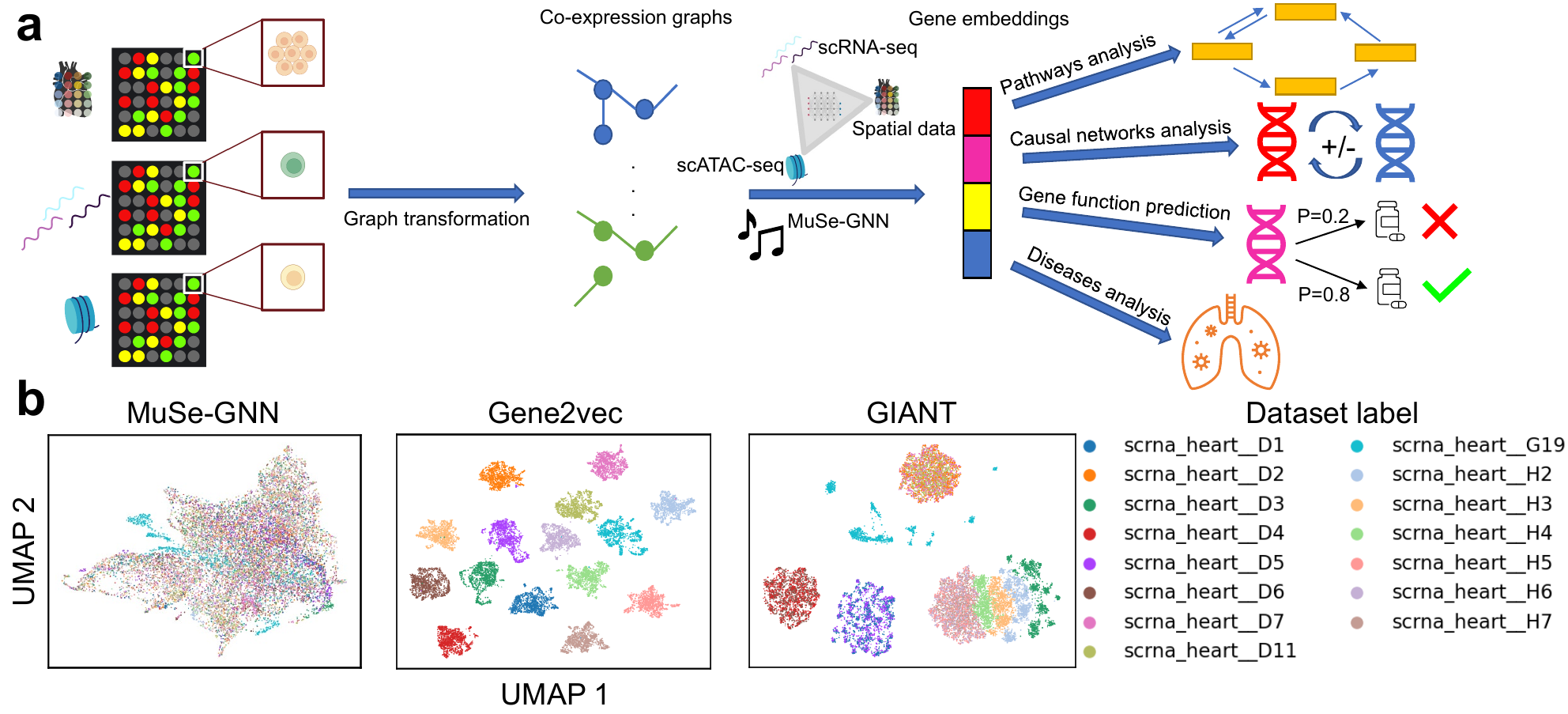}
    \caption{The workflow of MuSe-GNN, the visualization of gene embeddings, and the problems of two exitsing methods, GIANT and Gene2vec. \textbf{(a)} The process of learning gene embeddings by MuSe-GNN. Here we highlight the difference between single-cell data and spatial data, and the major applications of gene embeddings. Each spot from single-cell data represents one cell, while each spot from spatial data represents a mixture of cells. \textbf{(b)} UMAPs \cite{mcinnes2018umap} for the gene embeddings of MuSe-GNN, Gene2vec and GIANT. They are colored by datasets. Based on UMAPs we can conclude that both Gene2vec and GIANT failed to learn the gene similarity based on the datasets from the same tissue, while MuSe-GNN produces meaningful embeddings agnostic to datasets.
    \label{fig:problem data}}
\end{figure}

However, the traditional idea of multi-omics data integration known as using cells as anchors \cite{lin2022clustering, stuart2019comprehensive, hao2021integrated, butler2018integrating, dominguez2022cross, rosen2023towards} is only partially suitable because of the following two challenges. 1. Different omics data pose their own challenges. For example, the unit of observation of the spatial transcriptomic data is different from other single-cell data, as a single spatial location contains mixed information (see the left part of Figure \ref{fig:problem data} a) from various cells \cite{dong2021spatialdwls} and it is not appropriate to generate spatial clusters based on gene expression similarity \cite{chen2022unified}. Current research \cite{strober2019dynamic} also indicates that chromosome accessibility feature is not a powerful predictor for gene expression at the cell level. 2. The vast data volume in atlas-level studies challenges high-performance computing, risking out-of-memory or time-out errors \cite{lopez2018deep, wang2022respan}. With nearly 37.2 trillion cells in the human body \cite{rozenblatt2017human}, comprehensive analysis is computationally infeasible. 3. Batch effects may adversely impact analysis results \cite{leek2010tackling} by introducing noise. Consequently, an efficient and powerful method focusing on multi-omics and multi-tissue data (referred to as multimodal biological data) analysis is urgently needed to address these challenges. 

Acknowledging the difficulties arising from the cell-oriented viewpoint, prior work shifted focus to the gene perspective. Using gene sets as a summary of expression profiles, based on natural selection during the species evolution, may provide a more robust anchor \cite{argelaguet2021computational}. Protein-coding genes are also thought to interact with drugs \cite{cotto2018dgidb}, which are more relevant to diseases and drug discovery. Gene2vec \cite{du2019Gene2vec} is a method inspired by Word2vec \cite{church2017word2vec}, which learns gene representations by generating skip-gram pairs from the co-expression network. Recently, the Gene-based data Integration and ANalysis Technique (GIANT) \cite{chen2022unified} has been developed, based on Node2vec \cite{grover2016node2vec} and OhmNet \cite{zitnik2017predicting}, to learn gene representations from both single-cell and spatial datasets. However, as shown in Figure \ref{fig:problem data} (b), significant functional clustering for genes from different datasets but the same tissue was not observed based on the gene embeddings from these two models, because these methods do not infer the similarity of associated genes from different multimodal data. Additionally, they did not offer metrics to quantitatively evaluate the performance of gene embeddings compared to baseline models.

Here we introduce a novel model called \textbf{Mu}ltimodal \textbf{S}imilarity L\textbf{e}arning \textbf{G}raph \textbf{N}eural \textbf{N}etwork (MuSe-GNN) \footnote[1]{Codes of MuSe-GNN: \url{https://github.com/HelloWorldLTY/MuSe-GNN}} for multimodal biological data integration from a gene-centric perspective. The overall workflow of MuSe-GNN is depicted in Figure \ref{fig:problem data} (a). Figure \ref{fig:problem data} (b) shows MuSe-GNN's superior ability to learn functional similarity among genes across datasets by suitable model structure and novel loss function, comparing to GIANT and Gene2vec. MuSe-GNN utilizes weight-sharing Graph Neural Networks (GNNs) to encode genes from different modalities into a shared space regularized by the similarity learning strategy and the contrastive learning strategy. At the single-graph level, the design of graph neural networks ensures that MuSe-GNN can learn the neighbor information in each co-expression network, thus preserving the gene function similarity. At the cross-data level, the similarity learning strategy ensures that MuSe-GNN can integrate genes with similar functions into a common group, while the contrastive learning strategy helps distinguish genes with different functions. Furthermore, MuSe-GNN utilizes more robust co-expression networks for training and applies dimensionality reduction \cite{sugiyama2007dimensionality} to high-dimensional data \cite{palit2019meeting}. 

To the best of our knowledge, this is the first paper in gene representation learning that combines the Multimodal Machine Learning (MMML) concept \cite{baltruvsaitis2018multimodal, ektefaie2023multimodal} with deep GNNs \cite{kipf2016semi, shi2020masked} design. Both approaches are prevalent in state-of-the-art (SOTA) machine learning research, inspiring us to apply them to the joint analysis of large-scale multimodal biological datasets \footnote[2]{Download links: Appendix \ref{appendix: ds information}.}. As application examples, we first used the gene embeddings generated by MuSe-GNN to investigate crucial biological processes and causal networks for gene regulation in the human body. We also applied our model to analyze COVID and cancer datasets, aiming to unveil potential disease resistance mechanisms or complications based on specific differentially co-expressed genes. Lastly, we used gene embeddings from MuSe-GNN to improve the prediction accuracy for gene functions. Here genes are co-expressed means genes are connected with the same edge.

Given the lack of explicit metrics to evaluate gene embeddings, we proposed six metrics inspired by the batch effect correction problem \cite{leek2010tackling} in single-cell data analysis. We evaluated our model using real-world biological datasets from one technique, and the benchmarking results demonstrated MuSe-GNN's significant improvement from \textbf{20.1\%} to \textbf{97.5\%} in comprehensive assessment. To summarize the advantages of our model, MuSe-GNN addresses the outlined problem about cross-data gene similarity learning and offers four major contributions: 1. Providing an effective representation learning approach for multi-structure biological data. 2. Integrating genes from different omics and tissue data into a joint space while preserving biological information. 3. Identifying co-located genes with similar functions. 4. Inferring specialized causal networks of genes and the relationships between genes and biological pathways or between genes and diseases.

\section{Related work}
\textbf{Co-expression Network Analysis.} While direct attempts at joint analysis of gene functions across modalities are limited, there is relevant research in identifying correlation networks based on a single dataset. WGCNA \cite{langfelder2008wgcna} is a representative method that employs hierarchical clustering to identify gene modules with shared functions. However, as an early tool, WGCNA has restricted functionality. Its inference of co-expression networks necessitates more rigorous methods, and it cannot be directly applied to the analysis of multimodal biological data.

\textbf{Network Based Biological Representation Learning.} Apart from directly generating gene co-expression networks, learning quantitative gene representations in a lower dimensional space may better describe gene-gene relationships and facilitate downstream analyses. Gene2vec generates gene embeddings based on the co-expression network from a given database. However, it disregards the expression profile information and is based on the old Gene Expression Omnibus (GEO) up until 2017 \cite{du2019Gene2vec}. GIANT leverages Node2vec \cite{grover2016node2vec} and OhmNet \cite{zitnik2017predicting} to learn gene representations from both single-cell and spatial datasets by constructing graphs and hypergraphs for training. However, this approach still over-compresses multimodal biological datasets by removing expression profiles. Moreover, their co-expression networks created by Pearson correlation have a high false positive rate \cite{skinnider2019evaluating}. Additionally, some of the datasets used by GIANT are of low quality (see Appendix \ref{appendix: dataset}). There are also methods sharing a similar objective of learning embeddings from other datasets. GSS \cite{oh2022genomicsupersignature} aims to learn a set of general representations for all genes from bulk-seq datasets \cite{marguerat2010rna} using Principal Component Analysis (PCA) and clustering analysis. However, it is for bulk-seq data and cannot be directly applied to single-cell datasets from different tissues. Gemini \cite{woicik2023gemini} focuses on integrating different protein function networks, with graph nodes representing proteins rather than genes.

\textbf{Graph Transformer.} In the deep learning domain, Transformer \cite{vaswani2017attention, wang2020vd} is one of the most successful models, leveraging seq2seq structure design, multi-head self-attention design, and position encoding design. Many researchers have sought to incorporate the advantages of Transformer to graph structure learning. TransformConv \cite{shi2020masked} introduces a multi-head attention mechanism to the graph version of supervised classification tasks and achieved significant improvements. MAGNA \cite{wang2020multi} considers higher-level hop relationships in graph data to enhance node classification capabilities. Graphomer \cite{ying2021transformers} demonstrates the positive impact of the Transformer structure on various tasks using data from the Open Graph Benchmark Large-Scale Challenge \cite{hu2020open}, which is further extended by GraphomerGD \cite{zhang2023rethinking}. Recently, GPS \cite{rampavsek2022recipe} proposes a general Graph Transformer (GT) model by considering additional encoding types. Transformer architecture also contributes solutions to several biological questions \cite{zhang2023applications}. scBERT \cite{yang2022scbert} generates gene and cell embeddings using pre-training to improve the accuracy of cell type annotations. The substantial impact of these efforts highlights the crucial contribution of the Transformer architecture to graph data learning.

\section{Methods}
In the following sections of introducing MuSe-GNN, we will elaborate on our distinct approaches for graph construction that utilize multimodal biological data, followed by an explanation of our weight-sharing network architecture and the elements of the final loss function. 

\subsection{Preliminaries}
\textbf{GNN.} GNNs aim to learn the graph representation of nodes (features) for data with a graph structure. Modern GNNs iteratively update the representation of a node by aggregating the representations of its $k$-order neighbors ($k \geq 1$) and combining them with the current representation. As described in \cite{ying2021transformers}, considering a graph $G = \{V,E\}$ with nodes $V = \{v_1,v_2,...,v_n\}$, the AGGREGATE-COMBINE update for node $v_i$ is defined as:
\begin{equation}
a_i^{(l+1)}=\operatorname{AGGREGATE}^{(l+1)}\left(\left\{h_j^{(l)}: j \in \mathcal{N}\left(v_i\right)\right\}\right); h_i^{(l+1)}=\operatorname{COMBINE}^{(l+1)}\left(h_i^{(l)}, a_i^{(l+1)}\right),
\end{equation}
where $\mathcal{N}\left(v_i\right)$ represents the neighbors of node $v_i$ in the given graph, and $h_i^{(l)}$ and $h_i^{(l+1)}$ represent the node representation before and after updating, respectively.

\textbf{Problem Definition.} We address the gene embeddings generation task by handling multimodal biological datasets, denoted as $\mathcal{D}=\left(\left\{V_i, E_i\right\}\right)_{i=1}^{T}$. Our goal is to construct a model $\mathcal{M}(\cdot,\theta)$, designed to yield gene embeddings set $\mathcal{E} = \{e_1,...,e_T\} =\mathcal{M}(\mathcal{D}, \theta)$. In this context, $\mathcal{D}$ represents the input, $\theta$ represents the parameters, and $\mathcal{E}$ represents the output. In other words, we aim to harmonize gene information from diverse modalities within a unified projection space, thereby generating consolidated gene representations.

\subsection{Graph construction}
Before constructing gene graphs, our first contribution involves the selection of highly variable genes (HVGs) for each dataset. These HVGs constitute a group of genes with high variance that can represent the biological functions of given expression profiles. Moreover, considering co-expression networks is important for gene representation learning because it allows us to characterize gene-gene relationships. As sequencing depth, or the total counts of each cell, often serves as a confounding factor in the co-expression networks inference \cite{booeshaghi2022depth}, we employ two unique methodologies, scTransform \cite{hafemeister2019normalization} and CS-CORE \cite{su2022cell}, to process scRNA-seq and scATAC-seq data, thus creating gene expression profiles and co-expression networks unaffected by sequencing depth. For spatial transcriptomic data, our focus is on genes displaying spatial expression patterns. We use SPARK-X \cite{zhu2021spark} to identify such genes and then apply scTransform and CS-CORE. For detailed algorithmic information regarding these methods, please see Appendix \ref{appendix: method}. Additionally, we demonstrate the immunity of CS-CORE to batch effects when estimating gene-gene correlations in Appendix \ref{appendix: de ce analysis}. In all our generated graphs (equivalent to co-expression datasets), nodes represent \textbf{genes} and edges represent \textbf{co-expression} relation of genes.

\subsection{Cross-Graph Transformer}
To capitalize on the strengths of the Transformer model during our training process, we integrate a graph neural network featuring a multi-head self-attention design \cite{shi2020masked}, called TransformerConv, to incorporate co-expression information and generate gene embeddings. Details of TransformerConv can be found in Appendix \ref{Transformconv}. Incorporating multimodal information can estimate more accurate gene embeddings, supported by Appendix \ref{appendix: theory analysis of mmml}. The cross-graph transformer can efficiently learn gene embeddings containing gene functions across different graphs, advocated by the comparison of different network structure choices in Appendix \ref{appexp: ablation tests}.

\textbf{Weight sharing.} Given the variability among multimodal biological datasets, we employ a weight-sharing mechanism to ensure that our model learns shared information across different graphs, representing a novel approach for learning cross-graph relationships. We also highlight the importance of weight-sharing design in Appendix \ref{appexp: ablation tests}.
\begin{figure*}[ht]
    \centering
    \includegraphics[width=0.85\textwidth]{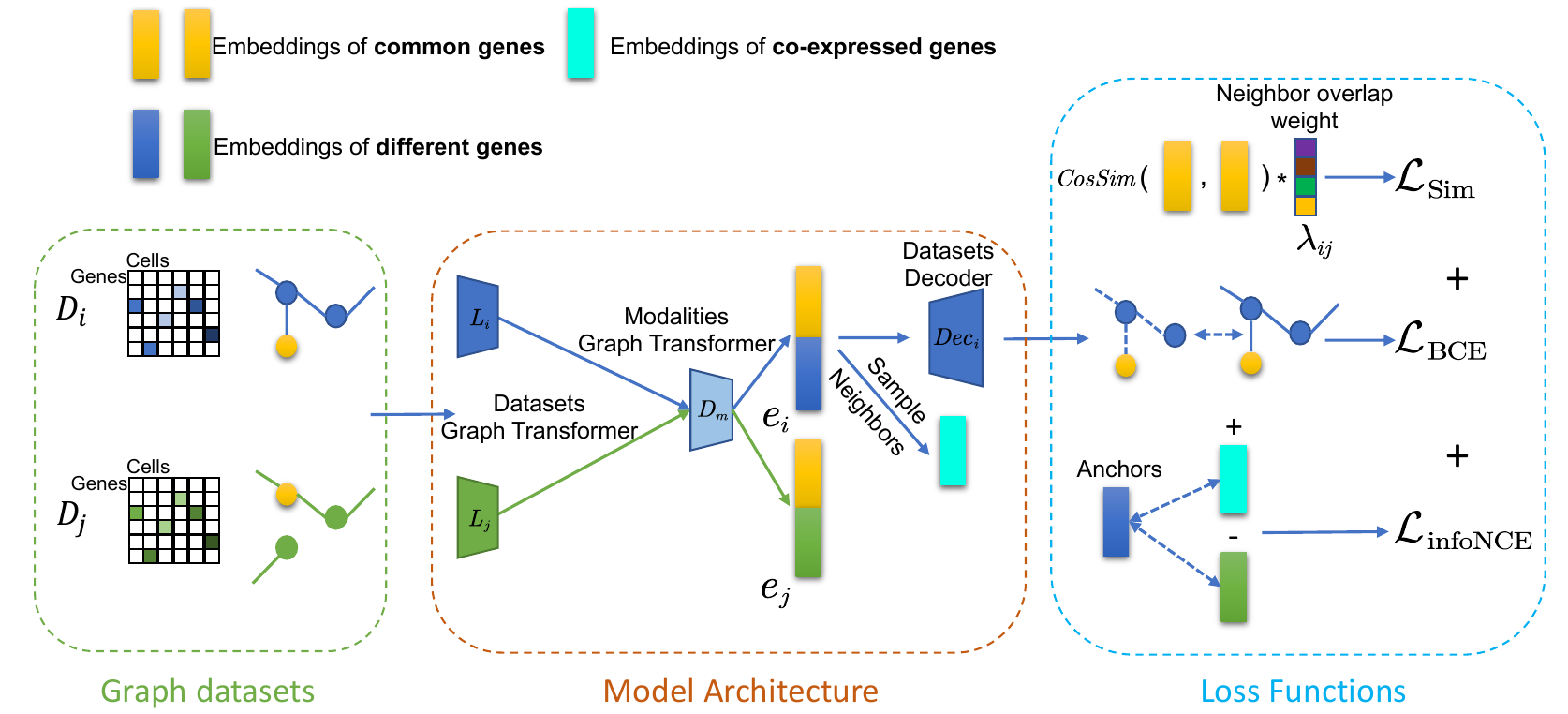}
    \caption{The overall model architecture and the design of loss functions for MuSe-GNN. The color of nodes in the green block represents common/different genes across two datasets. The brown block represents the network architecture of MuSe-GNN, and the blue block represents different loss function components of MuSe-GNN. The color gradients of the left two matrices represent different gene expression levels. 
    \label{fig:graph method}}
\end{figure*}

\textbf{Datasets \& Modalities Graph Transformer.} Drawing inspiration from the hard weight-sharing procedure \cite{caruana1993multitask, zhou2019auto}, we not only employ dataset-specific Graph Transformer (GT) layers $L_1, L_2, ..., L_n$ for each graph ($G_1, G_2, ..., G_n$) from the same modality $m$, but also connect all these dataset-specific layers to a set of shared GT layers, denoted as $D_m$. This design showcases our novel approach to incorporating weight-sharing into the GT framework. The forward process of MuSe-GNN, given dataset $i$ with network parameter $\theta_{*}$, is defined as follows:
\begin{equation}
X_i' = D_m(L_i(G_i; \theta_{L_i}); \theta_{D_m}).
\end{equation}
\textbf{Datasets Decoder.} Here we propose a dataset-specific decoder structure based on Multi-layer Perceptrons (MLP). This decoder model is crucial in reconstructing the co-expression relationships among different genes, showcasing our inventive use of MLP for this purpose. Given a graph $G_i$ and its corresponding gene embedding $e_i$, the decoding process of MuSe-GNN, with network parameter $\theta_{dec, i}$, is defined as follows:
\begin{equation}
E_{rec} = \textsc{DecoderMlp}(e_ie_i^T; \theta_{dec, i}),
\end{equation}
where $E_{rec}$ represents the reconstructed co-expression network.
\subsection{Graph Reconstruction Loss ($\mathcal{L}_{\text{BCE}}$)}
Within a single graph, we implement a loss function inspired by the Graph Auto-encoder (GAE) \cite{kipf2016variational}. This function is designed to preserve two key aspects: 1. the similarity among genes sharing common functions, and 2. the distinctions among genes with differing functions. This innovative use of a GAE-inspired loss function constitutes a significant contribution to the methodological design. For a graph $G_i = \{V_i, E_i\}$, the loss function for edge reconstruction is defined as:
\begin{equation}
\begin{aligned}
& e_i = \textsc{EncoderGNN}(\{V_i, E_i\}; \theta_{enc}) = D_m(L_i(G_i; \theta_{L_i}); \theta_{D_m}),  \\
& E_{rec} = \textsc{DecoderMlp}(e_ie_i^T; \theta_{dec, i}); E'_{rec} = \textsc{flatten}(E_{rec}); E'_i = \textsc{flatten}(E_{i}),\\
& \mathcal{L}_{\text{BCE}}= - \frac{1}{|E'_{rec}|} \sum_{t=1}^{|E'_{rec}|}\left[E'_{i}[t] \cdot \log E'_{rec}[t]+\left(1-E'_{i}[t]\right) \cdot \log \left(1-E'_{rec}[t]\right)\right],
\end{aligned}
\end{equation}
where $\textsc{EncoderGNN}$ and $\textsc{DecoderMlp}$ represent the encoder and decoder parts of our model, respectively. $\mathcal{L}_{\text{BCE}}$ denotes the computation of binary cross entropy loss (BCELoss) for the given input data. $E_{rec}$ is the reconstructed adjacency matrix and $|E'_{rec}|$ is the length of its flatten version. Further justification of the model design can be found in Appendix \ref{Extra model discussion}.
\subsection{Weighted-similarity Learning ($\mathcal{L}_{\text{Sim}}$)}
To integrate shared biological information across graphs from disparate datasets, we fuse the reconstruction loss of the input graph structure with a cosine similarity learning loss. In this process, we treat common HVGs between each pair of datasets as anchors. Our aim is to maximize the cosine similarity ($\textsc{CosSim}(\textbf{a},\textbf{b}) = \frac{\textbf{a} \cdot \textbf{b}}{||\textbf{a}||_2 \cdot ||\textbf{b}||_2}$) for each common HVG across two arbitrary datasets (represented by the yellow blocks in Figure \ref{fig:graph method}), utilizing their gene embeddings. However, in practice, different common HVGs may have different levels of functional similarity in two datasets, which is hard to be directly quantified. Consequently, we employ the shared community score as an indirect measurement, which is incorporated as a weight for the cosine similarity of different common HVG pairs within the final loss function. Considering two graphs $G_i = \{V_i, E_i\}$ and $G_j = \{V_j, E_j\}$, and one of their shared genes $g$, we identify the co-expressed genes of the given gene $g$ in both datasets, denoted as $N_{ig}$ and $N_{jg}$. Thus, the weight $\lambda_{ijg}$ for gene $g$ can be expressed as follows:
\begin{equation}
\lambda_{ijg} = \frac{|N_{ig} \cap N_{jg}|}{|N_{ig} \cup N_{jg}|}.
\end{equation}
We can iterate over all shared genes from 1 to $n$ between these two graphs, ultimately yielding a vector as $\lambda_{ij} = [\lambda_{ij1},..,\lambda_{ijn} ]$. This vector encapsulates the community similarity between the two graphs across all common HVGs. We can then modify the cosine similarity of various gene pairs by first multiplying this vector with cosine similarities and then summing the resultant values across all genes. The negation of the outcome is our final weighted similarity loss, denoted as $\mathcal{L}_{\text{Sim}}$. The ablation test for weighed similarity loss can be found in Appendix \ref{appexp: ablation tests}. More detailed explanations and evidence supporting this design in multimodal conditions can be found in Appendix \ref{appendix: commmon genes}.

\subsection{Self-supervised Graph Contrastive Learning ($\mathcal{L}_{\text{InfoNCE}}$)}
Specifically, when integrating multimodal biological data, we employ the contrastive learning strategy \cite{oord2018representation} to ensure that functionally similar genes are clustered together as closely as possible, while functionally different genes are separated apart from each other. We utilize Information Noise Contrastive Estimation (InfoNCE) as a part of our loss function to maximize the mutual information between the anchor genes and genes with the same functions. This loss is applicable to different genes in two arbitrary graphs during the training process. In general, if we represent the embeddings of $N$ genes as $\mathrm{Gene}_N = \{e_1,...,e_N\}$, InfoNCE is designed to minimize:
\begin{equation}
\label{equation infonce}
\mathcal{L}_{\text{InfoNCE}}=-\mathbb{E}\left[\log \frac{\exp \left(e \cdot k_{+} / \tau\right)}{\sum_{i=0}^K \exp \left(e \cdot k_i / \tau\right)}\right],
\end{equation}
where samples $\{k_0,k_1,k_2...\}$ compose a set of gene embeddings known as keys of one dictionary and $e$ is a query gene embedding. $k_{+}$ represents a positive sample of $e$ and $k_i$ denotes a negative sample of $e$. Equation \ref{equation infonce} can be interpreted as a log loss of a $(K+1)$-way Softmax classifier, which attempts to classify $e$ as $k_{+}$. $\tau$ is a temperature parameter. $\tau$ is set to 0.07 referenced in MoCo \cite{he2020momentum}.
\subsection{Final Loss Function}
In summary, the training objective of MuSe-GNN for graph $i$ comprises three components:
\begin{equation}
\begin{aligned}
\min_{e_i,e_j} & \text{ }  \mathcal{L}_{\text{BCE}}(\textsc{DecoderMlp}_i(e_ie_i^T), E_i) - \mathbb{E}\left[\textsc{CosSim}(e_i[\mathrm{Common}_{ij}], e_j[\mathrm{Common}_{ij}])\lambda_{ij}^T\right] \\
      &+ \lambda_c \mathcal{L}_{\text{InfoNCE}}(e_i[\mathrm{Diff}_i] \oplus e_j[\mathrm{Diff}_j],e_i[\mathrm{Diff}_{\mathcal{N}(i)}] \oplus e_j[\mathrm{Diff}_{\mathcal{N}(j)}]),
\end{aligned} 
\end{equation}
where $\mathrm{Common}_{ij}$ denotes the index set for common HVGs, $\mathrm{Diff}_i$ represents the index set for different HVGs in graph $i$, and $\mathrm{Diff}_{\mathcal{N}(i)}$ indicates the index set for neighbors of $\mathrm{Diff}_i$ in graph $i$. To expedite the training process and conserve memory usage, we sample graph $j$ for each graph $i$ during model training. We also employ multi-thread programming to accelerate the index set extraction process. $\lambda_c$ is the weight for InfoNCE loss.  All of the components in MuSe-GNN are supported by ablation experiments in Appendix \ref{appexp: ablation tests}. Details of hyper-parameter tuning can be found in Appendix \ref{appexp: hyper tuning}.

\section{Experiments}
\begin{algorithm}[htbp]
\caption{\textbf{Mu}ltimodal \textbf{S}imilarity L\textbf{e}arning \textbf{G}raph \textbf{N}eural \textbf{N}etwork (MuSe-GNN)}
\begin{algorithmic}[1]
\Statex \textbf{Input:} Model $\textsc{EncoderGNN}$, $\textsc{DecoderMlp}$; Dataset $\mathcal{D}=\left(\left\{V_i, E_i\right\}\right)_{i=1}^{T}$; Number of epochs $K$; Neighbor overlap list $\lambda$; Weight coefficient for contrastive learning $\lambda_{c}$;
\Statex \textbf{Helper Functions:} A function to find the genes with the same name $\textsc{FindCommonGenes}$;  A function to find the genes with different name $\textsc{FindDiffGenes}$; A function to calculate cosine similarity $\textsc{CosSim}$; A function to sample data from a graph collection $\textsc{Sample}$; A optimizer function to update weights $\textsc{Adam}$.
\Statex \textbf{Output:} Model $\textsc{EncoderGNN}$
\State INIT: initialize all parameters.
\For{$s$ in $K$ steps}
\For{$\text{id}, \{V, E\}$ in enumerate$(\mathcal{D})$}
\State $e \leftarrow \textsc{EncoderGNN}(\{V, E\}; \theta_{enc}); E_{rec} \leftarrow \textsc{DecoderMlp}(ee^T; \theta_{dec})$
\State $\mathcal{L} \leftarrow \mathcal{L}_{\text{BCE}}(E_{rec},E)$
\State $\text{id}_{\text{new}}, \{V_{\text{new}}, E_{\text{new}}\} \leftarrow \textsc{Sample}(G \backslash \{V, E\})$
\State $e_{\text{new}} \leftarrow \textsc{EncoderGNN}(\{V_{\text{new}}, E_{\text{new}}\};\theta_{enc})$
\State $e_{\text{Common}}, e'_{\text{Common}} \leftarrow \textsc{FindCommonGenes}(e, e_{\text{new}})$
\State $\mathcal{L} \leftarrow \mathcal{L} - \textsc{CosSim}(e_{\text{Common}}, e'_{\text{Common}})\lambda_{\text{id},\text{id}_{\text{new}}}^T$
\State $e_{\text{Diff}}, e'_{\text{Diff}}, e_{\text{Diff}_N}, e'_{\text{Diff}_N} \leftarrow \textsc{FindDiffGenes}(e, e_{\text{new}})$
\State $\mathcal{L} \leftarrow \mathcal{L} + \lambda_{c} \mathcal{L}_{\text{InfoNCE}}(e_{\text{Diff}} \oplus e_{\text{Diff}_N}, e'_{\text{Diff}} \oplus e'_{\text{Diff}_N})$
\State $\theta_{enc} \leftarrow \textsc{Adam}(\mathcal{L}, \theta_{enc}); \theta_{dec} \leftarrow \textsc{Adam}(\mathcal{L}, \theta_{dec})$
\EndFor
\EndFor
\State Return $\textsc{EncoderGNN}$
\end{algorithmic}
\end{algorithm}
\textbf{Datasets \& Embeddings generation.} Information on the different datasets used for different experiments is included at the beginning of each paragraph. The training algorithm of our model is outlined in Algorithm 1. We stored the gene embeddings in the AnnData structure provided by Scanpy \cite{wolf2018scanpy}. To identify groups of genes with similar functions, we applied the Leiden clustering algorithm \cite{traag2019louvain} to the obtained gene embeddings. Details can be found in Appendix \ref{appexp: function cluster identi}.

\textbf{Evaluation metrics.} For the benchmarking process, we used six metrics: edge AUC (AUC) \cite{pedregosa2011scikit}, common genes ASW (ASW) \cite{luecken2022benchmarking}, common genes graph connectivity (GC) \cite{luecken2022benchmarking}, common genes iLISI (iLISI) \cite{luecken2022benchmarking}, common genes ratio (CGR), and neighbors overlap (NO) to provide a comprehensive comparison. Detailed descriptions of these metrics can be found in Appendix \ref{appendix: metrics details}. We computed the metrics for the different methods and calculated the average rank (Avg Rank $\in [1,9]$). Moreover, to evaluate the model performance improvement, we applied min-max scaling to every metric across different models and computed the average score (Avg Score $\in [0,1]$).

\textbf{Baselines.} We selected eight models as competitors for MuSe-GNN. The first group of methods stems from previous work on learning embeddings for biomedical data, including Principal Component Analysis (PCA) \cite{pedregosa2011scikit} used by GSS, Gene2vec, GIANT (the SOTA model for gene representation learning), Weight-sharing Multi-layer Auto-encoder (WSMAE) used by \cite{yang2021multi} and scBERT (the SOTA model with pre-training for cell type annotation). The second group of methods comprises common unsupervised learning baseline models, including GAE, VGAE \cite{kipf2016variational} and Masked Auto-encoder (MAE) (the SOTA model for self-supervised learning) \cite{he2022masked}. MuSe-GNN, GIANT, GAE, VGAE, WSMAE and MAE have training parameter sizes between 52.5 and 349 M, and all models are tuned to their best performance. Details are shown in Appendix \ref{appexp: hyper tuning}. 

\textbf{Biological applications.} For the pathway analysis, we used Gene Ontology Enrichment Analysis (GOEA) \cite{goeacitation} to identify specific biological pathways enriched in distinct gene clusters with common functions. Moreover, we used Ingenuity Pathway Analysis (IPA) \cite{ipacitation} to extract biological information from the genes within various clusters, including causal networks \cite{kramer2014causal} and sets of diseases and biological functions. Biological pathways refer to processes identified based on the co-occurrence of genes within a particular cluster. The causal network depicts the relationships between regulatory genes and their target genes. Disease and biological function sets facilitate the discovery of key processes and complications associated with specific diseases. Using gene embeddings also improves the performance of models for gene function prediction. To visualize gene embeddings in a low-dimensional space, we utilized Uniform Manifold Approximation and Projection (UMAP) \cite{mcinnes2018umap}.
\begin{table}[htbp]
  \scriptsize
  \centering
  \caption{Avg Score across different tissues. Standard deviations are reported in Appendix \ref{appexp: benchmarking methods}.}
    \begin{tabular}{lrrrrrrrrrr}
    \toprule
    Methods & \multicolumn{1}{l}{Heart} & \multicolumn{1}{l}{Lung} & \multicolumn{1}{l}{Liver} & \multicolumn{1}{l}{Kidney} & \multicolumn{1}{l}{Thymus} & \multicolumn{1}{l}{Spleen} & \multicolumn{1}{l}{Pancreas} & \multicolumn{1}{l}{Cerebrum} & \multicolumn{1}{l}{Cerebellum} & \multicolumn{1}{l}{PBMC} \\
    \midrule
    PCA   & 0.52  & 0.48  & 0.56  & 0.47  & 0.56  & 0.60  & 0.51  & 0.62  & 0.53  & 0.51 \\
    Gene2vec & 0.40  & 0.37  & 0.33  & 0.29  & 0.21  & 0.31  & 0.24  & 0.27  & 0.31  & 0.19 \\
    GIANT & 0.50  & 0.40  & 0.33  & 0.38  & 0.58  & 0.33  & 0.56  & 0.29  & 0.28  & 0.28 \\
    WSMAE & 0.50  & 0.47  & 0.54  & 0.46  & 0.57  & 0.53  & 0.52  & 0.55  & 0.59  & 0.50 \\
    GAE   & 0.61  & 0.45  & 0.58  & 0.40  & 0.56  & 0.58  & 0.52  & 0.56  & 0.60  & 0.54 \\
    VGAE  & 0.64  & 0.32  & 0.33  & 0.38  & 0.56  & 0.31  & 0.33  & 0.41  & 0.33  & 0.47 \\
    MAE   & 0.36  & 0.47  & 0.50  & 0.45  & 0.41  & 0.52  & 0.39  & 0.50  & 0.49  & 0.50 \\
    scBERT & 0.41  & 0.49  & 0.55  & 0.62  & 0.17  & 0.58  & 0.46  & 0.60  & 0.61  & 0.58 \\
    MuSeGNN & \textbf{0.77} & \textbf{0.96} & \textbf{0.92} & \textbf{0.89} & \textbf{0.89} & \textbf{0.94} & \textbf{0.80} & \textbf{0.95} & \textbf{0.90} & \textbf{0.92} \\
    \bottomrule
    \end{tabular}%
  \label{benchmarking score}
\end{table}%
\subsection{Benchmarking Analysis}
We executed each method 10 times by using the same setting of seeds to show the statistical significance based on datasets across different tissues. The performance comparison of nine gene embedding methods is presented in Tables \ref{benchmarking score} and \ref{benchmarking table}. Based on these two tables, MuSe-GNN outperformed its competitors in terms of both average ranks and average scores across all the tissues. Based on Table \ref{benchmarking score}, for major tissues, such as heart and lung, MuSe-GNN's performance was 20.1\% higher than the second-best method and 97.5\% higher than the second-best method in heart and lung tissue, respectively. According to Appendix \ref{appexp: benchmarking methods}, MuSe-GNN's stability was also demonstrated through various metrics by comparing standard deviations, including AUC, GC, and NO. In contrast, methods such as Gene2vec, GAE, VGAE, MAE and scBERT exhibited significant instability in their evaluation results for kidney or thymus. Consequently, we concluded that MuSe-GNN is the best performing model for learning gene representation based on datasets from different tissues, making it applicable to learn gene embeddings from diverse multimodal biological data. 

\subsection{Analysis of Gene Embeddings from Multimodal Biological Data}
\begin{figure*}[ht]
    \centering
    \includegraphics[width=1.0\textwidth]{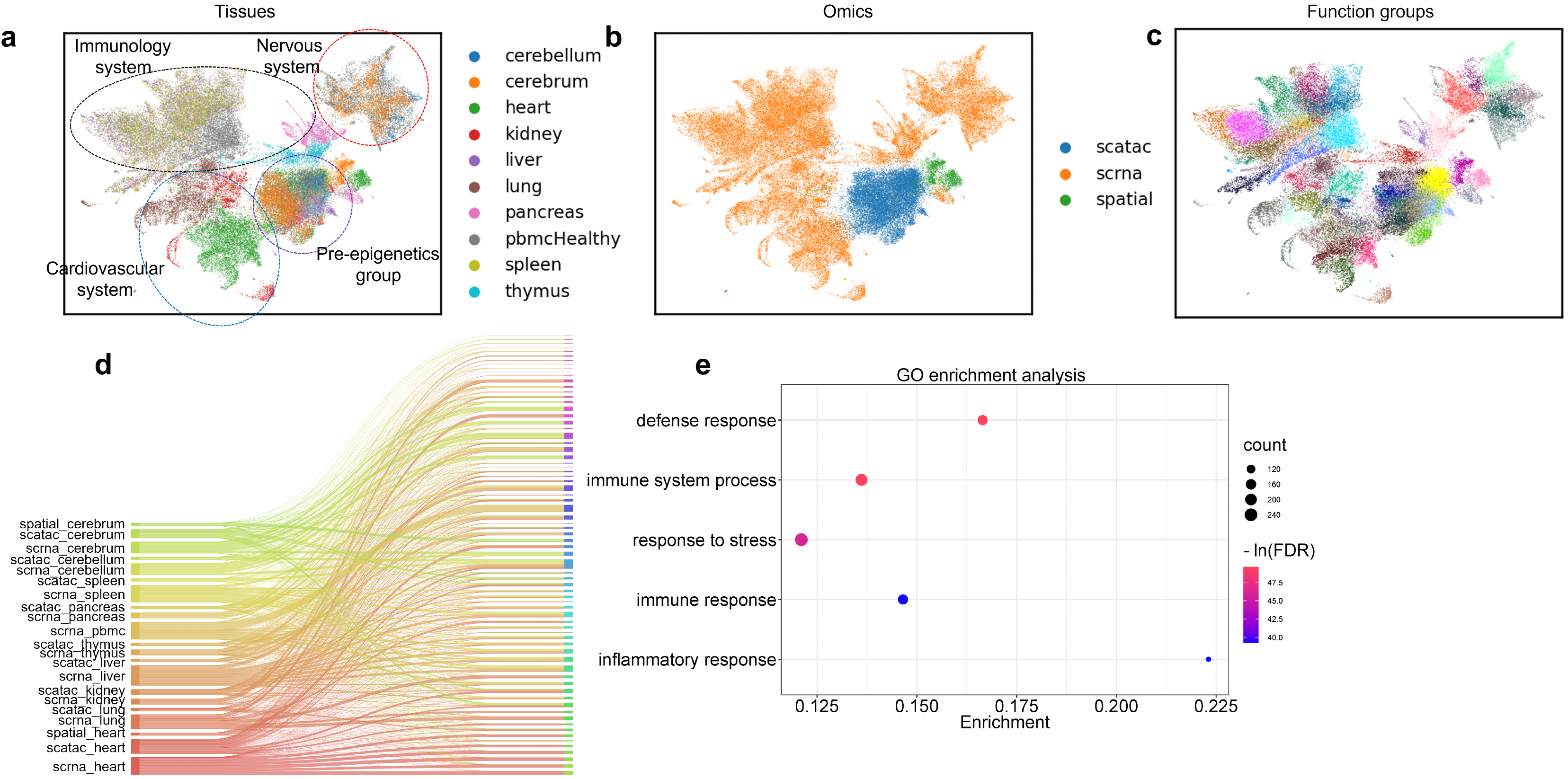}
    \caption{Gene representation learning results for multimodal biological data. \textbf{(a)} represents the UMAPs of gene embeddings colored by tissue type and highlighted by biological system. \textbf{(b)} represents the UMAPs of gene embeddings colored by omics type. \textbf{(c)} represents the UMAPs of gene embeddings colored by common function groups. \textbf{(d)} is a Sankey plot \cite{sankeyplotgithub} to show the overlap of different modalities in the same clusters. \textbf{(e)} shows the top5 pathways related to the genes in the special cluster discovered by GOEA. The bubble plots in this paper were created based on ggplot2 \cite{wickham2011ggplot2}. 
    \label{fig:multi tissue results}}
\end{figure*}

In Figure \ref{fig:multi tissue results}, we displayed the integration results for multimodal biological data from Humans. Figure \ref{fig:multi tissue results} (a) and (b) demonstrated that MuSe-GNN could successfully integrate genes from different modalities into a co-embedded space, allowing us to identify functional groups using the Leiden algorithm shown in Figure \ref{fig:multi tissue results} (c). Furthermore, Figure \ref{fig:multi tissue results} (d) revealed that most of the clusters in (c) were shared across different modalities. We also identified three significant functional groups in Figure \ref{fig:multi tissue results} (a): the nervous system (predominantly composed of the cerebrum and cerebellum \cite{durmaz2001multiple}), the cardiovascular system (mainly composed of heart, lung, and kidney \cite{fessler1997heart}), and the immunology system (primarily consisting of spleen, liver, and peripheral blood mononuclear cells (PBMC) \cite{nikzad2019human}). All systems are important in regulating the life activities of the body. We also uncovered a pre-epigenetics group (mainly consisting of scATAC-seq data without imprinting, modification, or editing), emphasizing the biological gap existing in multi-omics and the importance of post-transcriptional regulation.

Using GOEA, we could identify significant pathways enriched by different co-embedded gene clusters. For instance, Figure \ref{fig:multi tissue results} (d) displayed the top 5 pathways in an immunology system cluster. The rank was calculated based on the negative logarithm of the false discovery rate. Since all top pathways were related to immunological defense and response, it further supported the accuracy of our embeddings in representing gene functions. For our analysis of shared transcription factors and major pathways across all the tissues, please refer to Appendix \ref{appendix: tf analysis}. For our analysis of multi-species gene embeddings, please refer to Appendix \ref{appendix: multispecies}.
\begin{figure*}[ht]
    \centering
    \includegraphics[width=0.9\textwidth]{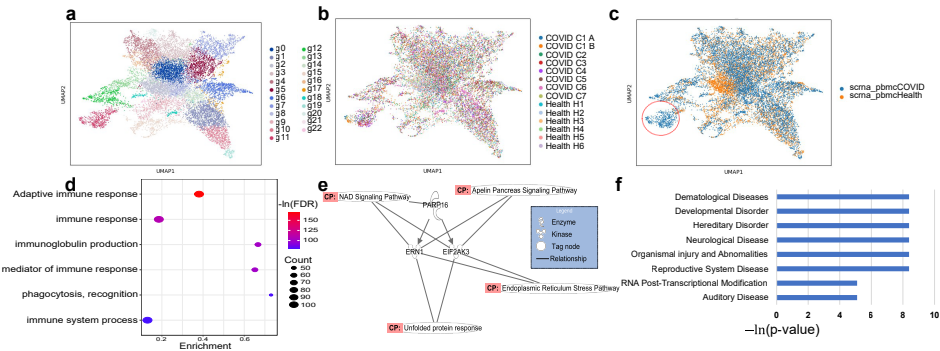}
    \caption{Gene embeddings from COVID samples and healthy samples. \textbf{(a)} represents the UMAPs of gene embeddings colored by functional groups. \textbf{(b)} represents the UMAPs of gene embeddings colored by datasets. \textbf{(c)} represents the gene embeddings colored by the conditions, and the red circle reflects the differential co-expression genes. \textbf{(d)} shows the top6 pathways related to the genes in the special cluster discovered by GOEA. \textbf{(e)} represents the causal network existing in the special cluster discovered by IPA. \textbf{(f)} represents the top diseases \& biological functions discovered by IPA.
    \label{fig:covid analysis}}
\end{figure*}
\subsection{Analysis of Gene Embeddings for Diseases}
We generated gene embeddings for human pancreas cells from samples with and without SARS-CoV-2 infection, as depicted in Figure \ref{fig:covid analysis} (a) and (b). We identified specific genes from COVID samples that did not align with control samples, which piqued our interest. These genes, highlighted by a red circle in Figure \ref{fig:covid analysis} (c), could be interpreted as differentially functional genes in diseased cells.

We conducted GOEA for the genes of interest and discovered a close relationship among these gene enrichment results and the top 5 pathways associated with immune activity. These results are displayed in Figure \ref{fig:covid analysis} (d). For the genes within our target cluster, we utilized IPA to identify the Entrez name of these genes, and 90.3\% (122/135) genes in our cluster are related to immunoglobulin. We could also infer the causal relationship existing in the gene regulatory activity of the immune system. For example, Figure \ref{fig:covid analysis} (e) showed a causal network inferred by IPA based on our genes cluster. PARP16, as an enzyme, can regulate ERN1 and EIF2AK3, and certain pathways are also related to this causal network. Moreover, we also showed the relation between the set of genes and Disease \& Bio functions in Figure \ref{fig:covid analysis} (f). We identified top related Diseases \& Bio functions ranked by negative logarithm of p-value, and all of these diseases could be interpreted as complications that may arise from new coronavirus infection \cite{aram2021covid,meral2022parental,salih2023hereditary,needham2020neurological,gayen2022covid,ardestani2021covid,srivastava2020role,sriwijitalai2020hearing}. Our extra analyses for lung cancer data can be found in Appendix \ref{appendix: lung cancer information}. 

\subsection{Analysis of Gene Embeddings for Gene Function Prediction.} Here we intend to predict the dosage-sensitivity of genes related to genetic diagnosis (as dosage-sensitive or not) \cite{theodoris2023transfer}. We used MuSe-GNN to generate gene embeddings for different datasets based on an unsupervised learning framework and utilized the gene embeddings as training dataset to predict the function of genes based on k-NN classifier. k-NN classifier is a very naive model and can reflect the contribution of gene embeddings in the prediction task.

In this task, we evaluated the performance of MuSe-GNN based on the dataset used in Geneformer \cite{theodoris2023transfer, franzen2019panglaodb}, comparing it to the prediction results based on raw data or Geneformer (total supervised learning). As shown in Table \ref{tab:gene-dosage pred analysis}, the prediction accuracy based on gene emebddings from MuSe-GNN is the highest one. Moreover, the performance of gene embeddings from MuSe-GNN is better than Geneformer, which is a totally supervised learning model. Such finding proves the advantages of MuSe-GNN in the application of gene function prediction task. Further application analysis can be found in Appendix \ref{appendix: gene function analysis}.

\begin{table}[ht]
  \centering
  \caption{Accuracy for dosage-sensitivity prediction}
    \begin{tabular}{lrrr}
    \toprule
          & \multicolumn{1}{l}{\textbf{MuSe-GNN (unsup)}} & \multicolumn{1}{l}{\textbf{Geneformer (sup)}} & \multicolumn{1}{l}{\textbf{Raw}} \\
    \midrule
    Accuracy & 0.77$\pm$0.01   & 0.74$\pm$0.06  & 0.75$\pm$0.01 \\
    \bottomrule
    \end{tabular}%
  \label{tab:gene-dosage pred analysis}%
\end{table}%

\section{Conclusion}
In this paper, we introduce MuSe-GNN, a model based on Multimodal Machine Learning and Deep Graph Neural Networks, for learning gene embeddings from multi-context sequencing profiles. Through experiments on various multimodal biological datasets, we demonstrate that MuSe-GNN outperforms current gene embedding learning models across different metrics and can effectively learn the functional similarity of genes across tissues and techniques. Moreover, we performed various biological analyses using the learned gene embeddings, leveraging the capabilities of GOEA and IPA, such as identifying significant pathways, detecting diseases, and inferring causal networks. Our model can also contribute to the study of the pathogenic mechanisms of diseases like COVID and lung cancer, and improve the prediction performance for gene functions. Overall, the gene representations learned by MuSe-GNN are highly versatile and can be applied to different analysis frameworks.

At present, MuSe-GNN does not accept graphs with nodes other than genes as input. In the future, we plan to explore more efficient approaches for training large models related to Multimodal Machine Learning and extend MuSe-GNN to a more general version capable of handling a broader range of multimodal biological data.

\section{Acknowledgements}
This research was supported in part by NIH R01 GM134005, R56 AG074015, and NSF grant DMS 1902903 to H.Z. We appreciate the comments, feedback, and suggestions from Chang Su, Zichun Xu, Xinning Shan, Yuhan Xie, Mingze Dong, and Maria Brbic.  

\bibliography{natbib}

\newpage

\appendix
\part{Appendix} 
\parttoc 

\section{GIANT's selection of datasets}
\label{appendix: dataset}

\begin{figure*}[ht]
    \centering
    \includegraphics[width=1\textwidth]{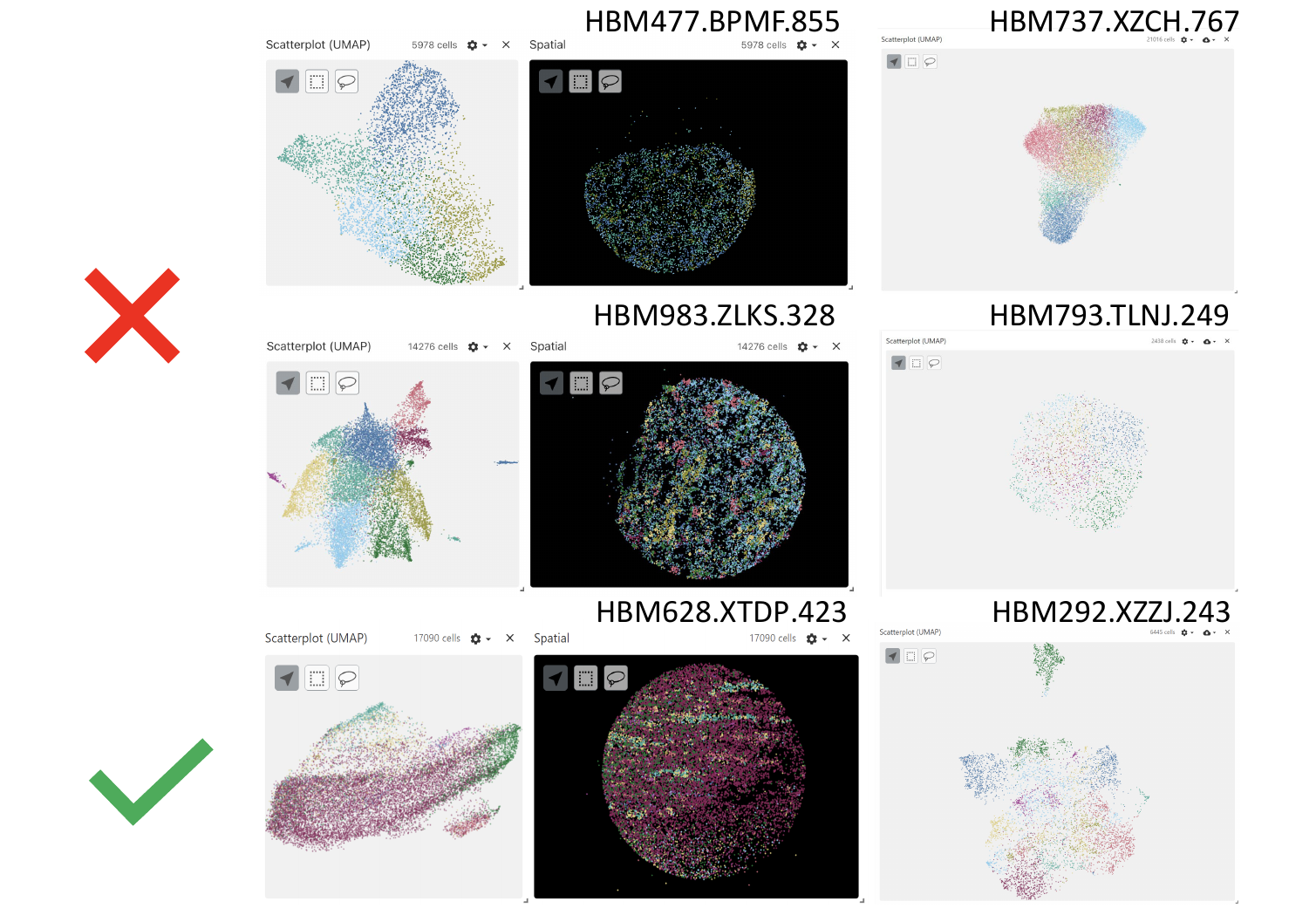}
    \caption{Problematic datasets selected by GIANT.
    \label{fig:problem dataset}}
\end{figure*}

The quality of datasets, as depicted in Figure \ref{fig:problem dataset}, impacts the conclusions derived from gene embedding learning methods. GIANT employed low-quality spatial transcriptomic and scATAC-seq data, which ideally should be circumvented at the research inception. The spatial data, HBM477.BPMF.855 and HBM983.ZLKS.328, exhibit considerable barcode expression measurement loss. Under normal circumstances, gene reads distribution in sectioned samples should form complete circles instead of truncated patterns. The scATAC-seq datasets, HBM737.XZCH.767 and HBM793.TLNJ.249, present challenges in discerning differences in gene expression assays across distinct cells, as evidenced by the UMAP visualization results. In an ideal scenario, cells marked with varying colors should occupy different positions in the low-dimensional representation. HBM628.XTDP.423 and HBM292.XZZJ.243 serve as examples of high-quality datasets.

\section{Constructing co-expression network and training datasets}
\label{appendix: method}
In this section, we outline the algorithmic details of preprocessing steps and network construction using CS-CORE, scTransform, and SPARK-X. Numerous databases exist for atlas-scale transcriptomic and epigenomic data analysis, including the Human BioMolecular Atlas Program (HuBMAP) \cite{hubmap2019human}, the Human Cell Atlas (HCA) \cite{rozenblatt2017human}, the 20 Cellular Senescence Network (SenNet) \cite{lee2022nih}, and more. We gather Single-cell RNA sequencing (scRNA-seq) from HCA datasets, \cite{cao2020human} and \cite{rosen2023towards}, Single-cell sequencing Assay for Transposase-accessible Chromatin (scATAC-seq) from HuBMAP datasets and \cite{cao2020human}, and high-quality spatial data from \cite{visium1, zhang2023spatial}. For distinct omics data, we implement one common and one specific process.

In the common process, we filter barcodes with gene expression counts below 200 and genes with expression counts below three in each barcode. We also filter Mitochondrial (MT) genes. For scRNA-seq datasets, we employ scTransform \cite{hafemeister2019normalization} to select HVGs and generate Pearson residuals, replacing raw expression with residuals. scTransform is the first model to incorporate sequencing depth as a covariate rather than directly applying the size factor to normalization. It eliminates confounding caused by sequencing depth in raw single-cell or spatial expression data, generating corrected gene expression profiles. These advantages make it a widely used normalization method \cite{booeshaghi2022depth}.

To construct scRNA-seq dataset co-expression networks based on Unique Molecular Identifier (UMI), we use CS-CORE \cite{su2022cell}, a state-of-the-art tool for co-expression inference based on UMI count data. CS-CORE demonstrates increased robustness and a lower false positive rate compared to other tools. For scATAC-seq datasets, we use Seurat \cite{hao2021integrated} to convert the original cells-peaks matrix into the cell-gene activity matrix, incorporating prior information. The cell-gene activity matrix can be processed similarly to the scRNA-seq data matrix, so subsequent preprocessing steps remain the same.

To construct spatial data co-expression networks, we consider spatial expression patterns (SE genes or spatially HVGs) and treat each barcode as a sample. We identify SE genes using SPARK-X \cite{zhu2021spark}, then generate corrected gene expression profiles based on scTransform and co-expression networks based on CS-CORE.
\subsection{CS-CORE}
\label{cscore}
Since our data are in UMI type, CS-CORE can be used to perform the co-expression network construction. Now considering we have $n$ cells and for one cell $i$, its expression profile can be denoted as a vector $(x_{i1}, ..., x_{ip})$, where $p$ is the number of genes. We can also use $s_i$ to represent the sequencing depth of cell $i$. Given the underlying expression levels from cell $i$ with $p$ genes as $\left(z_{i 1}, \ldots, z_{i p}\right)$, the assumption of CS-CORE for the expression profile follows:
\begin{equation}
\left(z_{i 1}, \ldots, z_{i p}\right) \sim F_p(\boldsymbol{\mu}, \boldsymbol{\Sigma}), \quad x_{i j} \mid z_{i j} \sim \operatorname{Poisson}\left(s_i z_{i j}\right)
\end{equation}
where $F_p$ is an unknown non-negative p-variate distribution with $\boldsymbol{\mu}$ as mean and $\boldsymbol{\Sigma}$ as covariance matrix. Here the observed expression $x_{ij}$ follows the Poisson measurement model depending on the underlying expression $z_{ij}$ and the sequencing depth $s_i=\sum_{j=1}^n x_{ij}$. CS-CORE applies a moment-based iteratively reweighted least squares (IRLS) estimation procedure to estimate the covariance matrix. Once we have the covariance matrix $\boldsymbol{\Sigma}_{p \times p}=[\sigma_{ij}]_{i=1...p}^{j=1...p}$, we can use the correlation $\rho_{jj'} = \frac{\sigma_{jj'}}{\sqrt{\sigma_{jj} \sigma_{j'j'}}}$ to estimate the co-expression relation between gene $j$ and gene $j'$.

With such an assumption, CS-CORE can model the relationship between gene $j$ and gene $j'$ based on a statistical test:
\begin{equation}
T_{j j^{\prime}}=\frac{\sum_i s_i^2\left(x_{i j}-s_i \mu_j\right)\left(x_{i j^{\prime}}-s_i \mu_{j^{\prime}}\right) g_{i j j^{\prime}}}{\sqrt{\sum_i s_i^4\left(s_i \mu_j+s_i^2 \sigma_{j j}\right)\left(s_i \mu_{j^{\prime}}+s_i^2 \sigma_{j^{\prime} j^{\prime}}\right) g_{i j j^{\prime}}^2}}
\end{equation}
The statistic $T_{jj'}$ under the null hypothesis assume gene $j$ and gene $j'$ are independent, which means there is no edge between these two genes. We chose to use p-value ($p<0.005$) to construct the edges in the co-expression network.
\subsection{scTransform}
\label{scTransform}
To remove the confounding effect of sequencing depth from the expression level, we first process the single-cell data using scTransform to get the Pearson residuals and then use the Pearson residuals as the initial embeddings of different genes. By assuming that the UMI count data follow the negative binomial distribution, for a given gene $g$ in cell $c$, we have:
\begin{equation}
\begin{aligned}
x_{g c} & \sim \mathrm{NB}\left(\mu_{g c}, \theta_g\right) \\
\ln \mu_{g c} & =\beta_{g 0}+\ln s_c,
\end{aligned}
\end{equation}

scTransform regularizes $\theta$ as a function of gene means $\mu$, by utilizing the Generalized Linear Model (GLM) with a log link function provided by the above equation. Furthermore, we can estimate the unknown parameters and calculate the Pearson residuals $Z_{gc}$ based on:
\begin{equation}
\begin{aligned}
& Z_{g c}=\frac{x_{g c}-\mu_{g c}}{\sigma_{g c}} \\
& \mu_{g c}=\exp (\beta_{g 0}+\ln s_c) \\
& \sigma_{g c}=\sqrt{\mu_{g c}+\frac{\mu_{g c}^2}{\theta_{g c}}}
\end{aligned}
\end{equation}
We can finally replace the original expression matrix with the residual matrix $Z$ generated by GLM, and store the expression matrix and the corresponding graph in Scanpy files.
\subsection{SPARK-X}
\label{SPARK-X}
The primary distinctions between single-cell transcriptomic data and spatial transcriptomic data involve two aspects: 1. In most spatially resolved data, each barcode represents a mixture of different cells. 2. The additional spatial information introduces spatial gene expression patterns (SE genes) for spatially resolved data. To identify SE genes in spatial transcriptomic data, SPARK-X employs a statistical test comparing the distance covariance matrix, which is constructed using barcode positions, and the expression covariance matrix, which is built from the gene expression profiles.

More specifically, for a spatial transcriptomic gene expression matrix with size $n\times d$, we can donate the matrix for coordinates of samples as $S = (s_1^T,...,s_n^T), s_i = (s_{i1}, s_{i2})$. Therefore, the whole expression matrix can be represented as : $y=(y_1(s_1),...,y_n(s_n))^T$. Our target is to test whether $y$ is independent from $S$, so we construct the expression covariance matrix based on $E=y(y^Ty)^{-1}y^T$, and the distance covariance matrix based on $\Sigma = S(S^TS)^{-1}S^T$. For the distance covariance matrix $S$, SPARK-X considers different kernels to describe different spatial expression patterns, including 1. Gaussian kernel $(s'_{i1}, s'_{i2}) = (exp(\frac{-s^2_{i1}}{2\sigma_1^2}), exp(\frac{-s^2_{i2}}{2\sigma_2^2}))$ and 2. Cosine kernel $(s'_{i1}, s'_{i2}) = (cos(\frac{2\pi s_{i1}}{\Phi_1}), cos(\frac{2\pi s_{i2}}{\Phi_2}))$. After centering these two matrices, we have $E_C$ and $\Sigma_C$. Therefore, the test statistic is:
\begin{equation}
T=\frac{\operatorname{trace}(E_{C} \Sigma_{C})}{n}
\end{equation}

This statistic follows a $\chi^2_1$ distribution and we record the p-value for each gene. The null hypothesis is that the gene expression is irrelevant to the position of barcodes. After finding the p-value list, we rank the p-value in ascending and select the top 1000 genes to perform normalization and co-expression graph construction. 

\begin{figure*}[ht]
    \centering
    \includegraphics[width=1\textwidth]{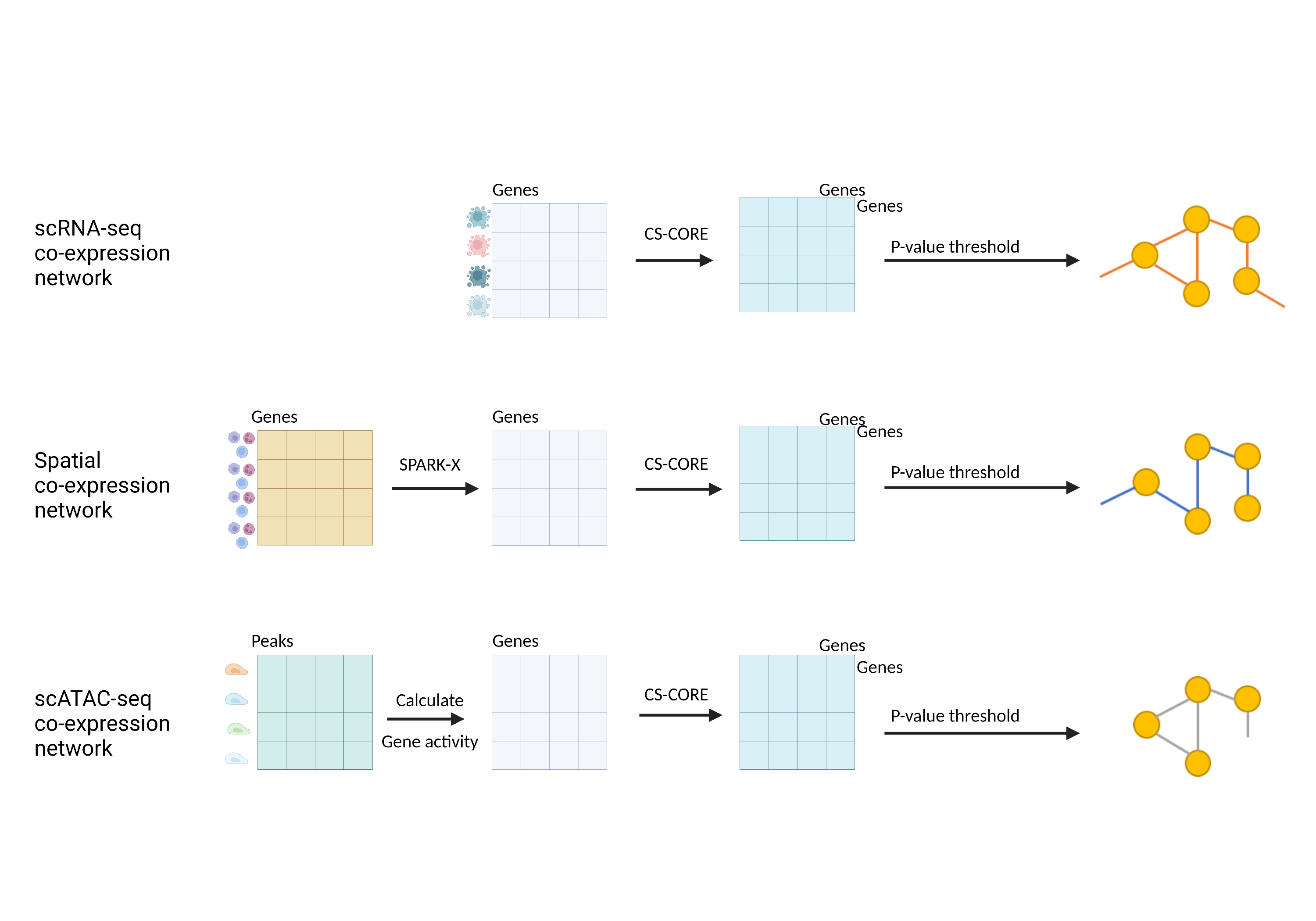}
    \caption{Schematic diagram of the graph data construction methods.
    \label{fig:graph overall}}
\end{figure*}

The whole process of the graph construction is shown in Figure \ref{fig:graph overall}. 

\subsection{Transformconv}
\label{Transformconv}
If we consider $c$ as the index of attention heads, the multi-head attention for GNN from node $j$ to node $i$ can be expressed as:
\begin{equation}
\begin{aligned}
q_{c, i}^{(l)} & =W_{c, q}^{(l)} h_i^{(l)}+b_{c, q}^{(l)}; k_{c, j}^{(l)} =W_{c, k}^{(l)} h_j^{(l)}+b_{c, k}^{(l)}, \\
e_{c, i j} & =W_{c, e} e_{i j}+b_{c, e}; \alpha_{c, i j}^{(l)} = \frac{\left\langle q_{c, i}^{(l)}, k_{c, j}^{(l)}+e_{c, i j}\right\rangle}{\sum_{u \in \mathcal{N}(i)}\left\langle q_{c, i}^{(l)}, k_{c, u}^{(l)}+e_{c, i u}\right\rangle},
\end{aligned}
\end{equation}
where $<q,k> = e^{\frac{q^Tk}{\sqrt{d}}}$. $\sqrt{d}$ is a scalar used to reduce gradient vanishing \cite{vaswani2017attention} as introduced in the original Transformer paper. The different vectors $q,k$ correspond to the query vector and the key vector, while $e$ represents the edge features. The attention $\alpha^{(l)}_{c, i j}$ denotes the $c_{th}$ attention value from node $j$ to $i$ for layer $l$. $h$ represents the node embedding, $W_{c, q}^{(l)}, W_{c, k}^{(l)}, W_{c, e}$ are weight matrices, and $b_{c, q}^{(l)}, b_{c, k}^{(l)}, b_{c, e}$ are bias terms.

We define the embedding of node $i$ in layer $l+1$ as $h_i^{l+1}$ and update the node embedding by:
\begin{equation}
\begin{aligned}
v_{c, j}^{(l)} & =W_{c, v}^{(l)} h_j^{(l)}+b_{c, v}^{(l)}, \\
h_i^{(l+1)} & =\|_{c=1}^C\left[\sum_{j \in \mathcal{N}(i)} \alpha_{c, i j}^{(l)}\left(v_{c, j}^{(l)}+e_{c, i j}\right)\right],
\end{aligned}
\end{equation}
where the vector $v$ represents the value vector. The operation $\|_{c=1}^C$ denotes the concatenation of $C$ total attention heads. $\mathcal{N}(i)$ represents the neighbors of node $i$. Additionally, we construct the residual link \cite{he2016deep} based on $h_i^{l-1}$ for $h_i^{l+1}$. The activation function employed to connect different layers is Mish ($\text{Mish}(x) = \tanh (\ln (1+\exp(x)))$) \cite{misra2019mish}. We also incorporated a GraphNorm layer \cite{cai2021graphnorm} to the connection path between hidden layers to enhance convergence.

\section{Differential co-expression analysis}
\label{appendix: de ce analysis}

In this section, we prove that using CS-CORE, we could obtain the co-expression relationships from multimodal biological data without the negative impact of confounding factors brought by the batch effect. 

\textbf{Theorem \ref{appendix: de ce analysis}.1} \textit{Given the true observed gene expression level $x$ and gene expression affected by batch effect $x'$, the construction of the co-expression network is the same based on the statistical test.}

\textit{Proof.} We consider the point $x'_{ij} = ax_{ij}+b$, where $a,b$ represent the effect caused by batch effect, and $x_{ij}$ represents the true biological data.  Moreover, based on the probabilistic model provided by CS-CORE, we have $\mathbb{E}\left(x_{i j}\right)=s_i \mu_j, Var(x_{ij}) = s_i \mu_j+s_i^2 \sigma_{j j} $. Therefore, for the two points $x_{i j}$ and $x'_{i j}$ we have:

\begin{equation}
\begin{aligned}
& x_{i j}=s_i \mu_j+\epsilon_{i j} \\
& \left(x_{i j}-s_i \mu_j\right)^2=s_i \mu_j+s_i^2 \sigma_{j j}+\eta_{i j} \\
& \left(x_{i j}-s_i \mu_j\right)\left(x_{i j^{\prime}}-s_i \mu_{j^{\prime}}\right)=s_i^2 \sigma_{j j^{\prime}}+\xi_{i j j^{\prime}}
\end{aligned}
\end{equation}

and

\begin{equation}
\begin{aligned}
& x'_{i j}=as_i\mu_j +b+a\epsilon_{i j} \\
& \left(x'_{i j}-(as_i\mu_j +b)\right)^2=a^2s_i\mu_i+s_i^2 a^2 \sigma_{j j}+\eta_{i j} \\
& \left(x'_{i j}-(as_i\mu_j +b)\right)\left(x'_{i j^{\prime}}-(as_i\mu_{j'} +b)\right)=s_i^2 a^2 \sigma_{j j^{\prime}}+\xi_{i j j^{\prime}}
\end{aligned}
\end{equation}

Based on this assumption, we can calculate the test statistic $T'_{j j^{\prime}}$ with batch effect, which is:
\begin{equation}
\begin{aligned}
T'_{j j^{\prime}}&=\frac{\sum_i s_i^2\left(x'_{i j}-a s_i \mu_j - b\right)\left(x'_{i j^{\prime}}-a s_i \mu_{j^{\prime}} - b\right) g_{i j j^{\prime}}}{\sqrt{\sum_i s_i^4\left(a^2 s_i \mu_j+a^2 s_i^2 \sigma_{j j}\right)\left(a^2 s_i \mu_{j^{\prime}}+a^2 s_i^2 \sigma_{j^{\prime} j^{\prime}}\right) g_{i j j^{\prime}}^2}} \\
&= \frac{a^2\sum_i s_i^2\left(x_{i j}-s_i \mu_j\right)\left(x_{i j^{\prime}}-s_i \mu_{j^{\prime}}\right) g_{i j j^{\prime}}}{a^2\sqrt{\sum_i s_i^4\left(s_i \mu_j+s_i^2 \sigma_{j j}\right)\left(s_i \mu_{j^{\prime}}+s_i^2 \sigma_{j^{\prime} j^{\prime}}\right) g_{i j j^{\prime}}^2}} \\
&= \frac{\sum_i s_i^2\left(x_{i j}-s_i \mu_j\right)\left(x_{i j^{\prime}}-s_i \mu_{j^{\prime}}\right) g_{i j j^{\prime}}}{\sqrt{\sum_i s_i^4\left(s_i \mu_j+s_i^2 \sigma_{j j}\right)\left(s_i \mu_{j^{\prime}}+s_i^2 \sigma_{j^{\prime} j^{\prime}}\right) g_{i j j^{\prime}}^2}} \\
&= T_{j j^{\prime}}
\end{aligned}
\end{equation}
This statistic is irrelevant to $a,b$. Therefore, the batch effect cannot affect our co-expression calculation, and our correlation defined by CS-CORE can reflect the correlation for true biological content.

\section{Theory analysis for multimodal machine learning}
\label{appendix: theory analysis of mmml}
Here we present a series of explanation for why using more modalities is more suitable in our task based on arguing that MMML method can generate more accurate estimate of the latent space representation for genes as the number of modalities increases.

Here we consider our datasets set $\mathcal{D}$ and different types of modalities $\mathcal{M}$ and $\mathcal{N}$, and their corresponding gene embeddings as $e_\mathcal{M}$ and  $e_\mathcal{N}$, where $\textsc{Card}(\mathcal{N})<\textsc{Card}(\mathcal{M})$. We also have the true representation $e^*$. Based on our model, we can estimate $\hat{e}_{\mathcal{M}}, \hat{e}_{\mathcal{N}}$. Here we intend to prove the theorem:

\textbf{Theorem \ref{appendix: theory analysis of mmml}.1} \textit{ $\hat{e}_{\mathcal{M}}$ is a better estimation for $e^*$ comparing to $\hat{e}_{\mathcal{N}}$ with $\textsc{Card}(\mathcal{N})<\textsc{Card}(\mathcal{M})$.}

Here we consider dataset $\mathcal{S} = \{(X_i, Y_i)\}_{i=1}^m$ as a general training dataset. In our specific cases, we can treat $X_i$ carries information with graph structure and $Y_i$ contains the edges information of $X_i$. Therefore, we have formalized the dataset we used in our unsupervised learning task. Our target is to minimize the empirical risk:
\begin{equation}
\begin{aligned} \min & \quad \hat{r}\left(h \circ g_{\mathcal{M}}\right) \triangleq \frac{1}{m} \sum_{i=1}^{m} \ell\left(h \circ g_{\mathcal{M}}\left(\mathbf{X}_{i}\right), Y_{i}\right) \\ \text { s.t. } & \quad h \in \mathcal{H}, g_{\mathcal{M}} \in \mathcal{G}_{\mathcal{M}}\end{aligned},
\end{equation}

where $l(\cdot , \cdot)$ represents the loss function, $h \circ g_{\mathcal{M}}\left(\mathbf{X}_{i}\right) = h(g_{\mathcal{M}}(\mathbf{X}_i))$ represents the composite function of $h$ and $g_{\mathcal{M}}$. Here function $g$ maps data from $X_i$ to the latent space, while function $h$ maps data  from latent space to the space of $Y_i$. We have $\mathcal{H}$ as the function class of $h$ and $G_{\mathcal{M}}$ is the function class of $g_{\mathcal{M}}$. We define the true map functions as $h^*$ and $g^*$. We donate $r(\cdot , \cdot)$ as risk.

To finish our proof, we need three assumptions. 

\textbf{Assumption \ref{appendix: theory analysis of mmml}.1.} \textit{The loss function is L-smooth with respect to the first ordinate, and is bounded by a constant C \cite{mohri2018foundations, tripuraneni2020theory, tripuraneni2021provable}.}

In our case, we have three components. BCELoss and Cosine Similarity follow this property \cite{xie2021genermodel, khromov2023some}. InfoNCE loss also follows this property proved by \cite{ma2021improving}.

\textbf{Assumption \ref{appendix: theory analysis of mmml}.2.} The true latent representation \(g^{\star}\) is contained in \(\mathcal{G}\), and the task mapping \(h^{\star}\) is contained in \(\mathcal{H}\) \cite{du2020few, tripuraneni2020theory, tripuraneni2021provable}.

\textbf{Assumption \ref{appendix: theory analysis of mmml}.3.} For any \(g^{\prime} \in \mathcal{G}^{\prime}\) and \(\mathcal{M} \subset[K], g^{\prime} \circ p_{\mathcal{M}}^{\prime} \in \mathcal{G}^{\prime}\) \cite{huang2021makes}.

Here the set $[K]$ represents modality set with $K$ modalities. $p_{\mathcal{M}}^{\prime}$ is a diagonal matrix using 1 for $ii$-th entry $\in \mathcal{M}$ and 0 otherwise.  

We also need one metric to evaluate the quality of our generated embeddings, so here we introduce our definition of \textit{latent representation quality}.

\textbf{Definition \ref{appendix: theory analysis of mmml}.1.} For any latent space generated by $g$, the \textbf{latent representation quality} is defined as \cite{huang2021makes} 
\begin{equation}
\eta(g)=\inf _{h \in \mathcal{H}}\left[r(h \circ g)-r\left(h^{*} \circ g^{*}\right)\right]
\end{equation}

One approach we can use to prove \textbf{Theorem \ref{appendix: theory analysis of mmml}.1} is to explore the order relationship between the upper bound of $\eta(g_{\mathcal{M}})$ and the upper bound of $\eta(g_{\mathcal{N}})$. That is, we intend to prove:
$$
\sup_{g_{\mathcal{M}} \in G_{\mathcal{M}}} \eta(g_{\mathcal{M}}) \leq \sup_{g_{\mathcal{N}} \in G_{\mathcal{N}}} \eta(g_{\mathcal{N}})
$$

To prove the above relation, we rely on a new theorem to compute the upper bound of $\eta(\cdot)$.

\textbf{Theorem \ref{appendix: theory analysis of mmml}.2.} \cite{huang2021makes} Let \(\mathcal{S}=\left\{\left(\mathbf{X}_{i}, Y_{i}\right)\right\}_{i=1}^{m}\) be a dataset of m examples drawn i.i.d. according to \(\mathcal{D}\). Let \(\mathcal{M}\)
be a subset of \([K]\). Assuming we have produced the empirical risk minimizers \(\left(\hat{h}_{\mathcal{M}}, \hat{g}_{\mathcal{M}}\right)\) training
with the \(\mathcal{M}\) modalities. Then, for all \(1>\delta>0\) with probability at least \(1-\delta:\)
$$
\eta\left(\hat{g}_{\mathcal{M}}\right) \leq 4 L \Re_{m}\left(\mathcal{H} \circ \mathcal{G}_{\mathcal{M}}\right)+4 L \Re_{m}(\mathcal{H} \circ \mathcal{G})+6 C \sqrt{\frac{2 \ln (2 / \delta)}{m}}+\hat{L}\left(\hat{h}_{\mathcal{M}} \circ \hat{g}_{\mathcal{M}}, \mathcal{S}\right),
$$
where \(\hat{L}\left(\hat{h}_{\mathcal{M}} \circ \hat{g}_{\mathcal{M}}, \mathcal{S}\right) \triangleq \hat{r}\left(\hat{h}_{\mathcal{M}} \circ \hat{g}_{\mathcal{M}}\right)-\hat{r}\left(h^{\star} \circ g^{\star}\right)\) is the centered empirical loss and $L$ represents the smoothness coefficient.

\textbf{Remark.} Now we consider sets $\mathcal{N} \subset \mathcal{M} \subset [K]$, and based on Assumption \ref{appendix: theory analysis of mmml}.3, we have $\mathcal{G}_\mathcal{N} \subset \mathcal{G}_\mathcal{M} \subset \mathcal{G}$. We can interpret this relation as the size of parameter space $\textsc{Param}$ follows the relation: $\textsc{Param}_\mathcal{N} \subset \textsc{Param}_\mathcal{M} \subset \textsc{Param}_\mathcal{G}$. Therefore, larger function class has a smaller empirical risk, so we have
\begin{equation}
\hat{L}\left(\hat{h}_{\mathcal{M}} \circ \hat{g}_{\mathcal{M}}, \mathcal{S}\right) \leq \hat{L}\left(\hat{h}_{\mathcal{N}} \circ \hat{g}_{\mathcal{N}}, \mathcal{S}\right)
\end{equation}

Based on Theorem \ref{appendix: theory analysis of mmml}.2, we can derive the upper bound of $\eta(\hat{g}_{\mathcal{N}})$ as:
$$
\eta\left(\hat{g}_{\mathcal{N}}\right) \leq 4 L \Re_{m}\left(\mathcal{H} \circ \mathcal{G}_{\mathcal{N}}\right)+4 L \Re_{m}(\mathcal{H} \circ \mathcal{G})+6 C \sqrt{\frac{2 \ln (2 / \delta)}{m}}+\hat{L}\left(\hat{h}_{\mathcal{N}} \circ \hat{g}_{\mathcal{N}}, \mathcal{S}\right),
$$
where $\Re_{m}(\cdot)$ represents the Rademacher complexity \cite{bartlett2002rademacher}. Here we can use the Right-hand Side (RHS) to approximate term $4 L \Re_{m}\left(\mathcal{H} \circ \mathcal{G}_{\mathcal{M}}\right) \sim \sqrt{C\left(\mathcal{H} \circ \mathcal{G}_{\mathcal{M}}\right) / m}$ and term $4 L \Re_{m}\left(\mathcal{H} \circ \mathcal{G}_{\mathcal{N}}\right) \sim \sqrt{C\left(\mathcal{H} \circ \mathcal{G}_{\mathcal{N}}\right) / m}$. Based on the basic structural property of Rademacher complexity \cite{bartlett2002rademacher}, we have 
$$
C\left(\mathcal{H} \circ \mathcal{G}_{\mathcal{N}}\right) \leq C\left(\mathcal{H} \circ \mathcal{G}_{\mathcal{M}}\right)
$$

Therefore, we can compute:
$$
\begin{aligned}
\eta\left(\hat{g}_{\mathcal{M}}\right) - \eta\left(\hat{g}_{\mathcal{N}}\right) & = 4 L \Re_{m}\left(\mathcal{H} \circ \mathcal{G}_{\mathcal{M}}\right) + \hat{L}\left(\hat{h}_{\mathcal{M}} \circ \hat{g}_{\mathcal{M}}, \mathcal{S}\right) - 4 L \Re_{m}\left(\mathcal{H} \circ \mathcal{G}_{\mathcal{N}}\right) - \hat{L}\left(\hat{h}_{\mathcal{N}} \circ \hat{g}_{\mathcal{N}}, \mathcal{S}\right) \\
&= \sqrt{\frac{C\left(\mathcal{H} \circ \mathcal{G}_{\mathcal{M}}\right)}{m}}-\sqrt{\frac{C\left(\mathcal{H} \circ \mathcal{G}_{\mathcal{N}}\right)}{m}} + \hat{L}\left(\hat{h}_{\mathcal{M}} \circ \hat{g}_{\mathcal{M}}, \mathcal{S}\right)-\hat{L}\left(\hat{h}_{\mathcal{N}} \circ \hat{g}_{\mathcal{N}}, \mathcal{S}\right) \\
& =\frac{\sqrt{C\left(\mathcal{H} \circ \mathcal{G}_{\mathcal{M}}\right)} - \sqrt{C\left(\mathcal{H} \circ \mathcal{G}_{\mathcal{N}}\right)}}{\sqrt{m}}
+ \hat{L}\left(\hat{h}_{\mathcal{M}} \circ \hat{g}_{\mathcal{M}}, \mathcal{S}\right)-\hat{L}\left(\hat{h}_{\mathcal{N}} \circ \hat{g}_{\mathcal{N}}, \mathcal{S}\right) \\
& \rightarrow  \hat{L}\left(\hat{h}_{\mathcal{M}} \circ \hat{g}_{\mathcal{M}}, \mathcal{S}\right)-\hat{L}\left(\hat{h}_{\mathcal{N}} \circ \hat{g}_{\mathcal{N}}, \mathcal{S}\right) |_{ m \rightarrow \infty} \\
& \leq 0
\end{aligned}
$$
\textbf{Remark.} Therefore, as the number of modalities increases, the representation quality of more modalities will be better than the representation quality of fewer modalities. Hence we proved Theorem \ref{appendix: theory analysis of mmml}.1 under large number of modalities cases. In the real biological application, genes can be active in many different scenarios, including different cells, different tissues and even different species. Therefore, our proof fits the context of MuSe-GNN's real application.

\section{Extra experiments, ablation test and model selection}
\label{appendix: experiments}

\subsection{Ablation tests}
\label{appexp: ablation tests}
We further investigate the contributions of various loss functions and GNN models to MuSe-GNN's performance. Our final loss function comprises three components: the graph reconstruction loss $\mathcal{L}_{\text{reconst}}$, the similarity maximization loss $\mathcal{L}_{\text{CosSim}}$, and the contrastive learning loss $\mathcal{L}_{\text{InfoNCE}}$. The reconstruction loss serves as the basic loss, and we assess our model's performance with different loss function components across all scRNA-seq datasets. The results are presented in Table \ref{table: ablation test} and \ref{table: ablation test score}.

\begin{table}[htbp]
\tiny
  \centering
  \caption{The overall benchmarking average rank for ablation tests based on different tissues.}
   \label{table: ablation test}
    \begin{tabular}{rrrrrrrrrrrrr}
    \toprule
    \multicolumn{1}{l}{$\mathcal{L}_{\text{CosSim}}$} & \multicolumn{1}{l}{$\mathcal{L}_{\text{InfoNCE}}$} & \multicolumn{1}{l}{Heart} & \multicolumn{1}{l}{Lung} & \multicolumn{1}{l}{Liver} & \multicolumn{1}{l}{Kidney} & \multicolumn{1}{l}{Thymus} & \multicolumn{1}{l}{Spleen} & \multicolumn{1}{l}{Pancreas} & \multicolumn{1}{l}{Cerebrum} & \multicolumn{1}{l}{Cerebellum} & \multicolumn{1}{l}{PBMC} & \multicolumn{1}{l}{Avg Rank} \\
    \midrule
          &       & 2.83  & 3.50  & 3.17  & 3.33  & 2.67  & 3.17  & 3.00  & 3.67  & 3.00  & \textbf{3.33}  & 3.17 \\
    \checkmark  &       & 2.17  & \textbf{1.67} & 2.00  & 2.17  & 2.50  & 2.00  & \textbf{1.50} & 2.00  & 2.17  & 5.67  & 2.38 \\
          & \checkmark    & 3.00  & 3.00  & 3.17  & 2.83  & 3.17  & 3.17  & 3.33  & 3.00  & 2.83  & 5.00  & 3.25 \\
    \checkmark     & \checkmark     & \textbf{2.00} & 1.83  & \textbf{1.67} & \textbf{1.67} & \textbf{1.67} & \textbf{1.67} & 2.17  & \textbf{1.33} & \textbf{2.00} & 3.83 & \textbf{1.98} \\
    \bottomrule
    \end{tabular}%
  
\end{table}%
\begin{table}[htbp]
\tiny
  \centering
  \caption{The overall benchmarking average score for ablation tests based on different tissues.}
   \label{table: ablation test score}
    \begin{tabular}{rrrrrrrrrrrrr}
    \toprule
    \multicolumn{1}{l}{$\mathcal{L}_{\text{CosSim}}$} & \multicolumn{1}{l}{$\mathcal{L}_{\text{InfoNCE}}$} & \multicolumn{1}{l}{Heart} & \multicolumn{1}{l}{Lung} & \multicolumn{1}{l}{Liver} & \multicolumn{1}{l}{Kidney} & \multicolumn{1}{l}{Thymus} & \multicolumn{1}{l}{Spleen} & \multicolumn{1}{l}{Pancreas} & \multicolumn{1}{l}{Cerebrum} & \multicolumn{1}{l}{Cerebellum} & \multicolumn{1}{l}{PBMC} & \multicolumn{1}{l}{Avg Score} \\
    \midrule
          &       & 0.42  & 0.13  & 0.21  & 0.24  & 0.36  & 0.19  & 0.28  & 0.09  & 0.37  & 0.23  & 0.25 \\
    \checkmark  &       & 0.65  & \textbf{0.81} & 0.77  & 0.75  & 0.57  & 0.70  & \textbf{0.83} & 0.71  & 0.68  & 0.66  & 0.71 \\
          & \checkmark     & 0.34  & 0.27  & 0.25  & 0.23  & 0.23  & 0.20  & 0.18  & 0.26  & 0.35  & 0.23  & 0.25 \\
    \checkmark     & \checkmark     & \textbf{0.78} & \textbf{0.81} & \textbf{0.94} & \textbf{0.80} & \textbf{0.70} & \textbf{0.96} & 0.53  & \textbf{0.91} & \textbf{0.74} & \textbf{0.90} & \textbf{0.81} \\
    \bottomrule
    \end{tabular}%
  
\end{table}%

Based on the results of this ablation test, we conclude that combining all three components to construct our loss function is the optimal choice, achieving a performance that is 224.0\% higher than the version without any additional regularization. Furthermore, using only contrastive learning may reduce the performance of our model, implying that learning similarity is more crucial than learning differences for the gene embedding generation task. When both components are included, the final average rank is 14.1\% higher than the version with only the similarity learning part. Thus, we need to incorporate both weighted similarity learning and contrastive learning in our final design to learn unified gene embeddings with improved performance.

Additionally, we compare the TransformConv framework with other GNN models, including GCN, Graph Attention Network (GAT) \cite{velivckovic2017graph}, SUGRL (GCN+MLP+Contrastive Learning) \cite{mo2022simple}, GPS \cite{rampavsek2022recipe}, GRACE (GCN+Contrastive Learning) \cite{zhu2020deep}, GraphSAGE \cite{hamilton2017inductive}, Graph Isomorphism Network (GIN) \cite{xu2018powerful}, GraphMAE \cite{hou2022graphmae} and Graphormer \cite{ying2021transformers}. We set the number of epochs to 1000 for this comparison. The results are displayed in Table \ref{Table: model selection}.

\begin{table}[htbp]
  \centering
  \small
  \caption{The overall benchmarking score and rank for model selection.}
  \label{Table: model selection}
    \begin{tabular}{lrrrrrrrr}
    \toprule
    Methods & \multicolumn{1}{l}{ASW} & \multicolumn{1}{l}{AUC} & \multicolumn{1}{l}{iLISI} & \multicolumn{1}{l}{GC} & \multicolumn{1}{l}{CGR} & \multicolumn{1}{l}{NO} & \multicolumn{1}{l}{Avg Rank} & \multicolumn{1}{l}{Avg Score} \\
    \midrule
    GCN   & 0.74$\pm$0.02  & 0.82$\pm$0.01  & 0.32$\pm$0.02  & 0.44$\pm$0.02  & 0.32$\pm$0.04  & 0.31$\pm$0.01  & 3.5 & 0.51 \\
    GAT   & 0.74$\pm$0.09  & \textbf{0.83$\pm$0.01}  & 0.14$\pm$0.03  & 0.42$\pm$0.03  & 0.00$\pm$0.00  & 0.15$\pm$0.03  & 5 & 0.17 \\
    SUGRL & \textbf{0.88$\pm$0.01}  & 0.62$\pm$0.00  & 0.45$\pm$0.01  & 0.43$\pm$0.03  & 0.38$\pm$0.02  & 0.29$\pm$0.00  & 3.5 & 0.56 \\
    GPS & 0.77$\pm$0.02  & 0.63$\pm$0.00  & 0.50$\pm$0.01  & \textbf{0.72$\pm$0.03}  & 0.57$\pm$0.02  & 0.30$\pm$0.01  & 2.67 & 0.68 \\
    GRACE & 0.82$\pm$0.02  & 0.81$\pm$0.00  & 0.35$\pm$0.01  & 0.42$\pm$0.04  & 0.11$\pm$0.01  & 0.25$\pm$0.01  & 4.17 & 0.48 \\
    GraphSAGE & \multicolumn{1}{l}{OOM} & \multicolumn{1}{l}{OOM} & \multicolumn{1}{l}{OOM} & \multicolumn{1}{l}{OOM} & \multicolumn{1}{l}{OOM} & \multicolumn{1}{l}{OOM} & 7.00 & 0 \\
    GIN   & \multicolumn{1}{l}{OOM} & \multicolumn{1}{l}{OOM} & \multicolumn{1}{l}{OOM} & \multicolumn{1}{l}{OOM} & \multicolumn{1}{l}{OOM} & \multicolumn{1}{l}{OOM} & 7.00 & 0 \\
    GraphMAE   & \multicolumn{1}{l}{OOM} & \multicolumn{1}{l}{OOM} & \multicolumn{1}{l}{OOM} & \multicolumn{1}{l}{OOM} & \multicolumn{1}{l}{OOM} & \multicolumn{1}{l}{OOM} & 7.00 & 0 \\
    Graphormer   & \multicolumn{1}{l}{OOM} & \multicolumn{1}{l}{OOM} & \multicolumn{1}{l}{OOM} & \multicolumn{1}{l}{OOM} & \multicolumn{1}{l}{OOM} & \multicolumn{1}{l}{OOM} & 7.00 & 0 \\
    Transformconv & 0.75$\pm$0.01  & 0.78$\pm$0.02  & \textbf{0.51$\pm$0.02}  & 0.68$\pm$0.04  & \textbf{0.61$\pm$0.02}  & \textbf{0.31$\pm$0.00}  & \textbf{2.17} & \textbf{0.80} \\
    \bottomrule
    \end{tabular}%
  
\end{table}%

\begin{figure}[htbp]
    \centering
    \includegraphics[width=1\textwidth]{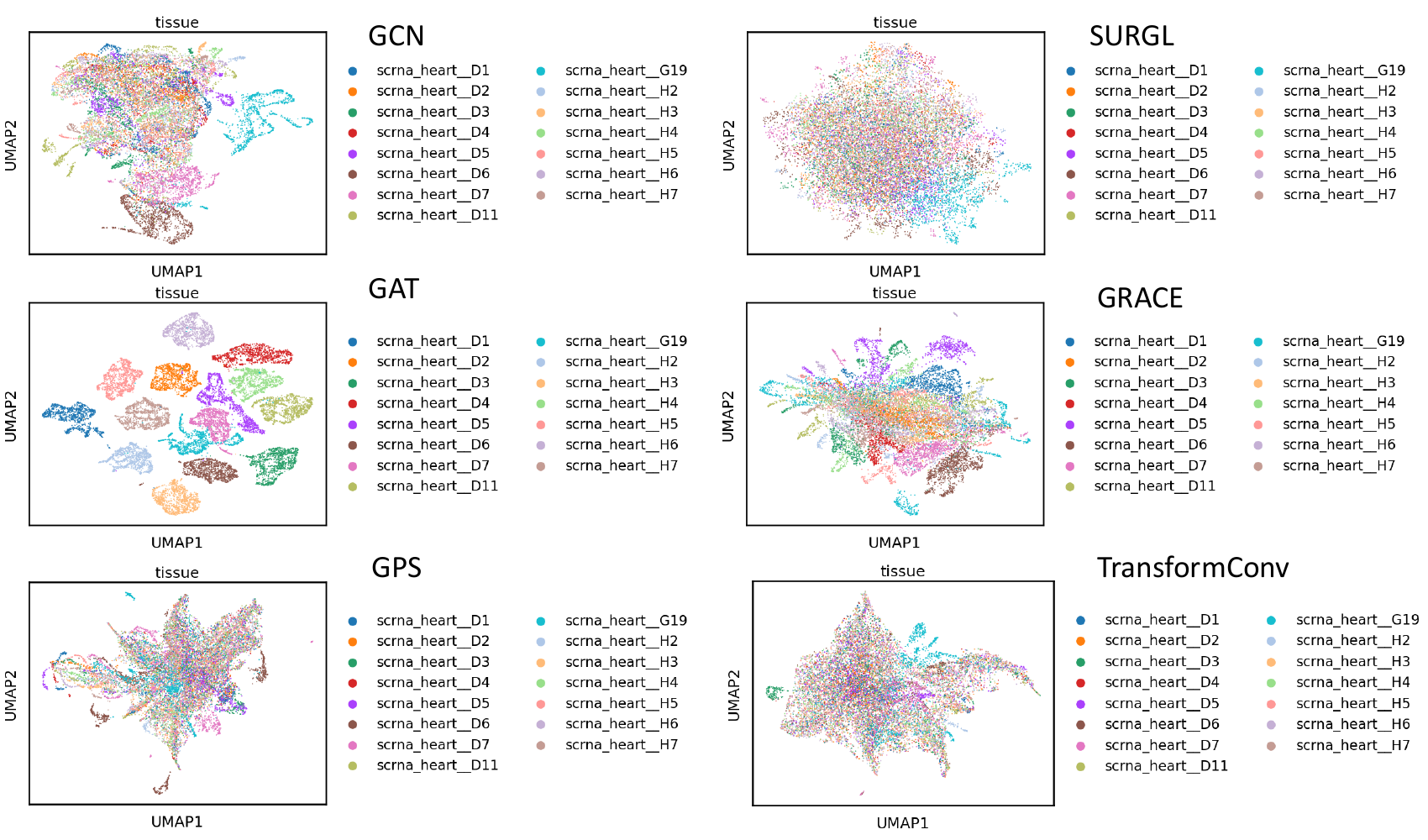}
    \caption{Gene embeddings based on different GNNs.
    \label{fig:network structure}}
\end{figure}

In this table, OOM means out of memory. We can conclude that using TransformConv as the basic graph neural network framework is the best choice, which is 17.6\% higher than the version based on GPS (the top2 model). Moreover, based on Figure \ref{fig:network structure}, we can also discover that methods based on GCN or GAT failed to learn the gene function similarity across different datasets. Therefore, choosing TransformConv is reasonable, and the major contribution of Transformer model towards gene embedding learning task is the multi-head attention design.

We also conduct experiments to justify the need for weight-sharing design. In this comparison, we evaluate the performance and model statistics of two models (with weight-sharing design (ws) and without weight-sharing design (w/o ws)). As shown in Table \ref{table: overall rank for all tissues}, the combined high or equivalent rank of weight-sharing designs in comparison to w/o weight-sharing designs is nearly consistent across all tissue types. Furthermore, the weight-sharing design can also reduce the model's training cost by decreasing the model size, the number of trainable parameters, and the total training time, as confirmed in Table \ref{table: model statistics}.

\begin{table}[htbp]
  \centering
  \small
  \caption{Comparison between with weight-sharing design and without weight-sharing design.}
  \label{table: overall rank for all tissues}
    \begin{tabular}{lrrrrrrrrrrr}
    \toprule
    Methods & \multicolumn{1}{l}{Heart} & \multicolumn{1}{l}{Lung} & \multicolumn{1}{l}{Liver} & \multicolumn{1}{l}{Kidney} & \multicolumn{1}{l}{Thymus} & \multicolumn{1}{l}{Spleen} & \multicolumn{1}{l}{Pancreas} & \multicolumn{1}{l}{Cerebrum} & \multicolumn{1}{l}{Cerebellum} & \multicolumn{1}{l}{PBMC} & \multicolumn{1}{l}{Avg Rank} \\
    \midrule
    ws    & 1.50  & \textbf{1.33} & \textbf{1.33} & \textbf{1.33} & \textbf{1.33} & \textbf{1.33} & 1.67  & 1.50  & \textbf{1.17} & \textbf{1.17} & \textbf{1.37} \\
    w/o ws  & 1.50  & 1.67  & 1.67  & 1.67  & 1.67  & 1.67  & \textbf{1.33} & 1.50  & 1.83  & 1.83  & 1.63 \\
    \bottomrule
    \end{tabular}%
\end{table}%

\begin{table}[htbp]
  \centering
  \caption{Model statistics.}
  \label{table: model statistics}
    \begin{tabular}{lrrr}
    \toprule
    Methods & \multicolumn{1}{l}{model size (MB)} & \multicolumn{1}{l}{\# of parameters (M)} & \multicolumn{1}{l}{training time per epoch (s)} \\
    \midrule
    ws    & \textbf{1333}  & \textbf{349} & \textbf{3.64} \\
    w/o ws & 1346.8 & 353 & 4.85 \\
    \bottomrule
    \end{tabular}%
\end{table}%

\begin{figure}[ht]
    \centering
    \includegraphics[width=1\textwidth]{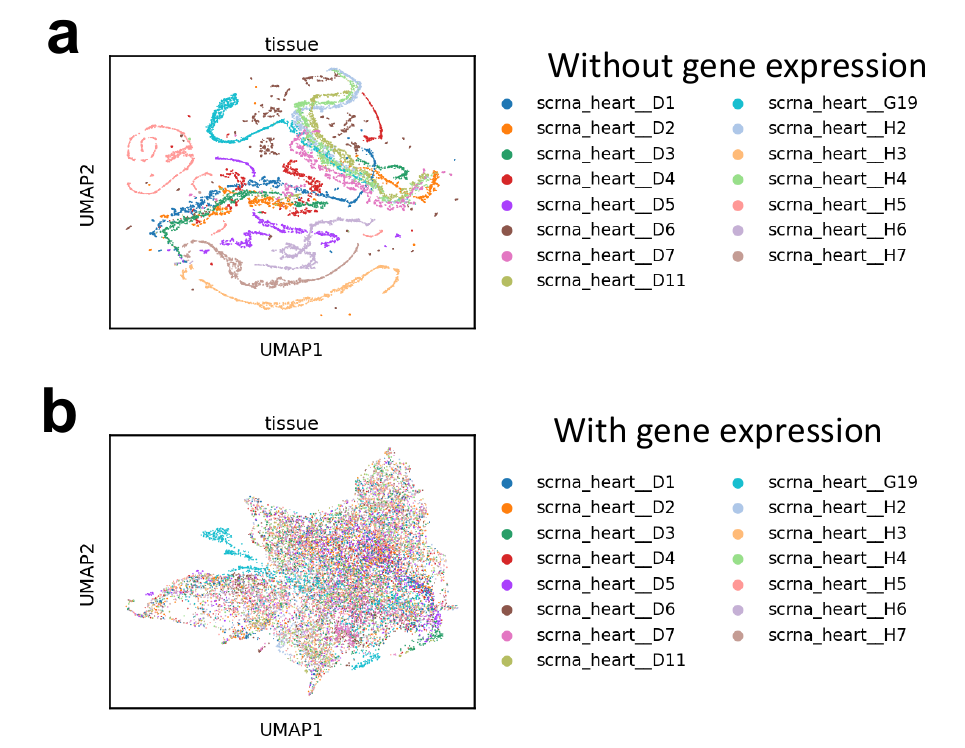}
    \caption{Gene embeddings based on different input information. \textbf{(a)} represents the UMAPs of gene embeddings without expression profiles and \textbf{(b)} represents the UMAPs of gene emebddings with expression profiles.
    \label{fig:information input}}
\end{figure}

To demonstrate the necessity of both expression profiles and graph structure for learning gene embeddings with biological information, we design an experiment in which all nodes in different graphs are replaced with the same features while maintaining the original graph structure. Based on the results shown in Figure \ref{fig:information input} and Table \ref{table:include feature or not}, we can conclude that without the information from expression profiles, it becomes difficult for our model to learn unified gene embeddings because of the decline in scores of different metrics.
\begin{table}[htbp]
  \centering
  \caption{Comparison between with feature design and without feature design.}
    \begin{tabular}{lrrrrrr}
    \toprule
    Methods & \multicolumn{1}{l}{ASW} & \multicolumn{1}{l}{AUC} & \multicolumn{1}{l}{iLISI} & \multicolumn{1}{l}{GC} & \multicolumn{1}{l}{CGR} & \multicolumn{1}{l}{NO} \\
    \midrule
    with feature & 0.75$\pm$0.01  & \textbf{0.78$\pm$0.02}  & \textbf{0.53$\pm$0.01}  & \textbf{0.73$\pm$0.04}  & \textbf{0.65$\pm$0.02}  & \textbf{0.31$\pm$0.00} \\
    w/o feature & 0.75$\pm$0.05  & 0.64$\pm$0.02  & 0.24$\pm$0.04  & 0.43$\pm$0.03  & 0.02$\pm$0.01  & 0.10$\pm$0.03 \\
    \bottomrule
    \end{tabular}%
  \label{table:include feature or not}%
\end{table}%

To demonstrate the necessity of Weighted Similarity Learning (WSL) design, we design an experiment to analyze the performance of different model structures with WSL or w/o WSL. According to Table \ref{table:weighted similarity learning}, we found that MuSe-GNN with WSL performed better than MuSe-GNN without WSL across all of the metrics in heart tissue. Therefore, we need to utilize WSL manchism to effectively learn the gene-gene relationships across different datasets.

\begin{table}[htbp]
  \centering
  \caption{Comparison between with weighted similarity learning design and without weighted similarity learning design.}
    \begin{tabular}{lrrrrrr}
    \toprule
    Methods & \multicolumn{1}{l}{ASW} & \multicolumn{1}{l}{AUC} & \multicolumn{1}{l}{iLISI} & \multicolumn{1}{l}{GC} & \multicolumn{1}{l}{CGR} & \multicolumn{1}{l}{NO} \\
    \midrule
    with WSL & \textbf{0.75$\pm$0.01}  & \textbf{0.78$\pm$0.02}  & \textbf{0.53$\pm$0.01}  & \textbf{0.73$\pm$0.04}  & \textbf{0.65$\pm$0.02}  & \textbf{0.31$\pm$0.00} \\
    w/o WSL & 0.69$\pm$0.02  & 0.66$\pm$0.03  & 0.46$\pm$0.03  & 0.57$\pm$0.07  & 0.48$\pm$0.05  & 0.29$\pm$0.00 \\
    \bottomrule
    \end{tabular}%
  \label{table:weighted similarity learning}%
\end{table}%

\subsection{Hyper-parameter tuning}
\label{appexp: hyper tuning}

Hyper-parameter settings for various models can be divided into two groups: 1. Methods such as PCA, Gene2vec, GIANT, WSMAE, GAE, VGAE, and MAE, which are free of pre-training, are configured with the best hyper-parameters based on parameter searching to ensure fairness. Details are included in Table \ref{tab:hyper parameters for all models}. 2. Method such as scBERT uses the pre-trained model. The optimal hyper-parameters for MuSe-GNN are provided in Table \ref{Table: hyperparameter result}. Number of trainable parameters for different models is summarized in Table \ref{tab:number of trainable parameters}.

\begin{table}[htbp]
  \centering
  \caption{Hyper-parameter candidates for benchmarking methods.}
    \begin{tabular}{ll}
    \toprule
    Models & Hyper-parameters \\
    \midrule
    PCA   & Dim:[32,128] \\
    Gene2vec & Dim:[32,128] \\
    GIANT & Dim:[32,128] \\
    WSMAE & LR:[1e-4,1e-2]; Batch size:[1000,3000]; Dim:[32,128] \\
    GAE   & LR:[1e-4,1e-2]; Batch size:[1000,3000]; Dim:[32,128] \\
    VGAE  & LR:[1e-4,1e-2]; Batch size:[1000,3000]; Dim:[32,128]  \\
    MAE   & LR:[1e-4,1e-2]; Batch size:[1000,3000]; Dim:[32,128]  \\
    \bottomrule
    \end{tabular}%
  \label{tab:hyper parameters for all models}%
\end{table}%

\begin{table}[ht]
\caption{Hyper-parameters list for MuSe-GNN.}
\label{Table: hyperparameter result}
\centering
\begin{tabular}{|l|l|lll}
\cline{1-2}
Parameter         & Value &  &  &  \\ \cline{1-2}
Epoch             & 2000  &  &  &  \\ \cline{1-2}
LR of encoder     & 1e-4  &  &  &  \\ \cline{1-2}
LR of decoder     & 1e-3  &  &  &  \\ \cline{1-2}
$\lambda_c$       & 1e-2  &  &  &  \\ \cline{1-2}
Dim of embeddings & 32    &  &  &  \\ \cline{1-2}
Sample size & 100    &  &  &  \\ \cline{1-2}
\end{tabular}
\end{table}

\begin{table}[htbp]
  \centering
  \caption{Number of trainable parameters for different models}
    \begin{tabular}{lr}
    \toprule
    Models & \multicolumn{1}{l}{\# of parameters (M)} \\
    \midrule
    MuSe-GNN & 349 \\
    GIANT & 260 \\
    VGAE  & 210 \\
    WSMAE & 105 \\
    GAE   & 52.5 \\
    MAE   & 52.5 \\
    \bottomrule
    \end{tabular}%
  \label{tab:number of trainable parameters}%
\end{table}%

\begin{figure}[ht]
    \centering
    \includegraphics[width=1\textwidth]{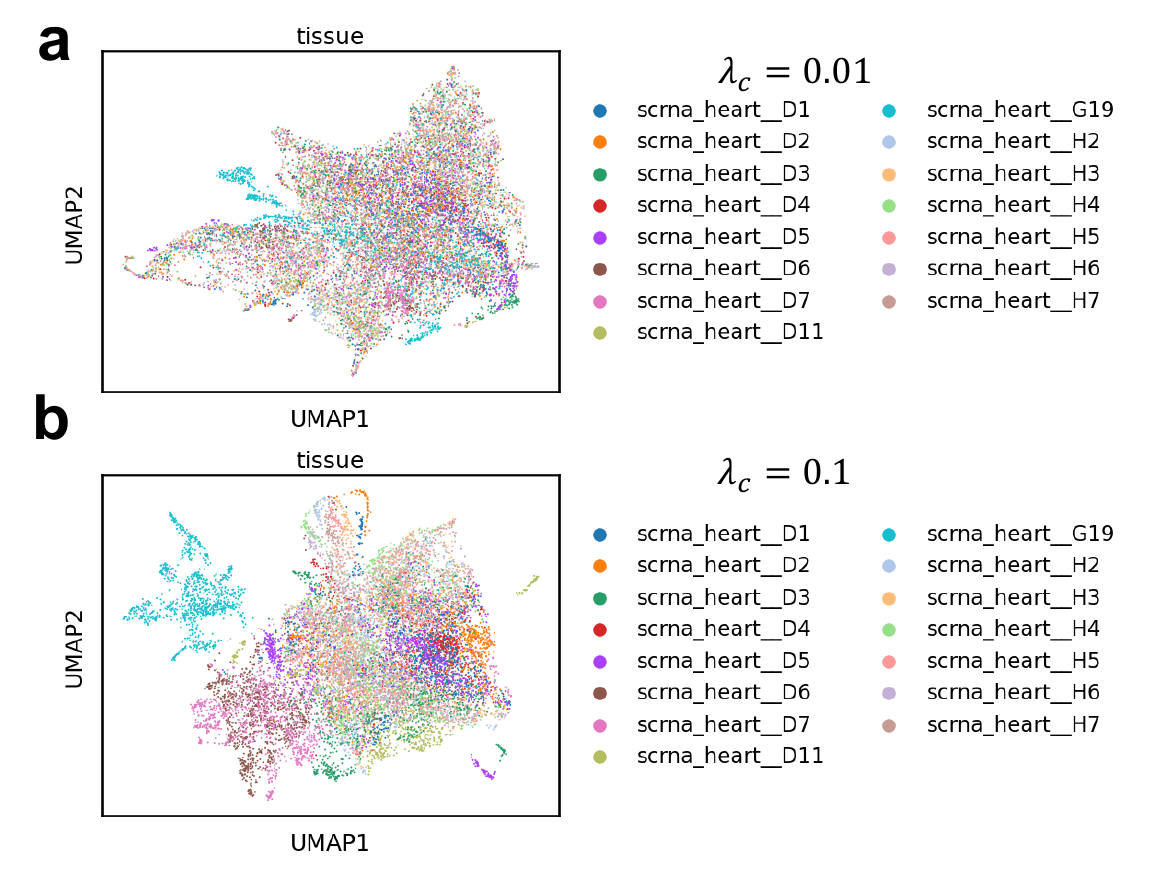}
    \caption{Gene embeddings on different $\lambda_c$.
    \label{fig:lambdac plot}}
\end{figure}

The average rank results for various hyper-parameters are presented in Table \ref{Table: hyperparameter selection}. We set the initial parameter settings as follows: Epoch = 2000, LR of encoder = 1e-2, LR of decoder = 1e-3, $\lambda_c$ = 1e-2, Dim = 32, and Sample size (for contrastive learning) = 100. Notably, for the value choice of $\lambda_c$, we have two candidates with the same average rank result. To further compare the gene embeddings generated by models based on these two choices, we plot the UMAPs for these two models. According to Figure \ref{fig:lambdac plot}, we observe that choosing $\lambda_c = 0.1$ fails to integrate genes from dataset G19. Therefore, we ultimately set $\lambda_c$ = 1e-2.

Additionally, the dimensions of gene embeddings should not be set to a very large value, not only due to the results in our table but also because of theoretical constraints and practical feasibility. Generally, an excessively large number of embedding settings may cause the number of hidden layer neurons to exceed the number of features in the input partial dataset, increasing the probability of overfitting by introducing more noise \cite{liu2008optimized}. Furthermore, a larger number of dimensions will consume more computational resources, potentially leading to out-of-memory issues, such as when setting the Dim of embeddings = 256.

\begin{table}[htbp]
  \centering
  \caption{Hyper-parameter searching results.}
    \label{Table: hyperparameter selection}
    \begin{tabular}{|c|c|c|c|}
    \toprule
    LR for encoder & Avg Rank & InfoNCE penalty & Avg Rank \\
    \midrule
    0.01  & 3.17  & 1     & 3.5 \\
    0.001 & 2.50  & 0.1   & \textbf{2.13} \\
    0.0005 & 2.33  & 0.01  & \textbf{2.13} \\
    0.0001 & \textbf{1.83} & 0.001 & 2.25 \\
    \midrule
    LR for decoder & Avg Rank & Dim   & Avg Rank \\
    \midrule
    0.01  & 3.43  & 32    & \textbf{1.86} \\
    0.001 & \textbf{1.86} & 64    & 2.00 \\
    0.0005 & 2.29  & 128   & 2.14 \\
    0.0001 & 2.43  & 256   & 4.00 \\
    \midrule
    batch size & Avg Rank & sample size & Avg Rank \\
    \midrule
    1000  & 3     & 50    & 2.38 \\
    1500  & 3     & 100   & \textbf{1.75} \\
    2000  & \textbf{1.5} & 150   & 2.88 \\
    3000  & 2.5   & 200   & 2.5 \\
    \bottomrule
    \end{tabular}%
  
\end{table}%

\subsection{Details about benchmarking experiments}
\label{appexp: benchmarking methods}

The average ranks result for all the scRNA-seq datasets is displayed in Table \ref{benchmarking table}. The experimental results for all the scRNA-seq datasets are displayed in Tables \ref{Table: heart result} to \ref{Table: pbmc result} (unscaled). Based on the benchmark results for different tissues, we can conclude that MuSe-GNN consistently outperforms other methods according to specific metrics, such as iLISI and NO. Furthermore, the gene embeddings learned by MuSe-GNN are relatively more stable than those from other methods, as evidenced by the analysis of standard deviation. Considering the overall performance, MuSe-GNN, as the top-performing method, also surpasses the second-best methods from 20.1\% to 97.5\% and surpasses the third-best methods from 26.9\% to 99.7\%.

\begin{table}[htbp]
  \scriptsize
  \centering
  \caption{The average ranks for gene embeddings benchmarking across different tissues.}
  \label{benchmarking table}
    \begin{tabular}{lrrrrrrrrrr}
    \toprule
    Methods & \multicolumn{1}{l}{Heart} & \multicolumn{1}{l}{Lung} & \multicolumn{1}{l}{Liver} & \multicolumn{1}{l}{Kidney} & \multicolumn{1}{l}{Thymus} & \multicolumn{1}{l}{Spleen} & \multicolumn{1}{l}{Pancreas} & \multicolumn{1}{l}{Cerebrum} & \multicolumn{1}{l}{Cerebellum} & \multicolumn{1}{l}{PBMC} \\
    \midrule
    PCA   & 4.33  & 4.33  & 4.67  & 5.50  & 4.00  & 4.17  & 4.83  & 4.50  & 5.50  & 4.00 \\
    Gene2vec & 5.67  & 6.67  & 6.50  & 6.67  & 7.33  & 7.17  & 6.83  & 6.67  & 6.00  & 7.33 \\
    GIANT & 6.00  & 4.67  & 5.33  & 4.67  & 3.67  & 5.67  & 3.83  & 5.50  & 6.00  & 6.17 \\
    WSMAE & 5.83  & 5.33  & 4.83  & 4.83  & 3.83  & 5.17  & 5.00  & 5.33  & 4.33  & 5.00 \\
    GAE   & 4.33  & 5.83  & 5.33  & 5.50  & 5.17  & 5.33  & 4.50  & 6.00  & 5.00  & 5.33 \\
    VGAE  & 3.50  & 6.33  & 6.33  & 6.67  & 4.83  & 6.50  & 6.17  & 6.67  & 6.83  & 5.83 \\
    MAE   & 7.00  & 5.17  & 5.17  & 5.17  & 5.50  & 5.17  & 5.67  & 5.33  & 5.33  & 5.33 \\
    scBERT & 5.67  & 4.50  & 4.33  & 3.17  & 7.83  & 3.83  & 4.83  & 3.17  & 3.17  & 3.67 \\
    MuSeGNN & \textbf{2.67} & \textbf{2.17} & \textbf{2.50} & \textbf{2.83} & \textbf{2.67} & \textbf{2.00} & \textbf{2.83} & \textbf{1.67} & \textbf{2.67} & \textbf{2.33} \\
    \bottomrule
    \end{tabular}%
\end{table}%


\begin{table}[htbp]
  \centering
  \caption{Benchmark score table for Heart.}
  \label{Table: heart result}
    \begin{tabular}{lrrrrrrr}
    \toprule
    Methods & \multicolumn{1}{l}{ASW} & \multicolumn{1}{l}{AUC} & \multicolumn{1}{l}{iLISI} & \multicolumn{1}{l}{GC} & \multicolumn{1}{l}{CGR} & \multicolumn{1}{l}{NO}  \\
    \midrule
    PCA   & 0.77$\pm$0.01  & 0.64$\pm$0.00  & 0.43$\pm$0.00  & 0.49$\pm$0.02  & 0.30$\pm$0.01  & 0.27$\pm$0.00   \\
    Gene2vec & 0.82$\pm$0.03  & 0.76$\pm$0.00  & 0.09$\pm$0.01  & 0.65$\pm$0.03  & 0.00$\pm$0.00  & 0.10$\pm$0.02   \\
    GIANT & \textbf{0.84$\pm$0.01}  & 0.55$\pm$0.00  & 0.26$\pm$0.02  & 0.48$\pm$0.03  & 0.10$\pm$0.01  & 0.01$\pm$0.00   \\
    WSMAE & 0.79$\pm$0.01   & 0.72$\pm$0.00  & 0.34$\pm$0.01  & 0.43$\pm$0.04  & 0.11$\pm$0.02  & 0.26$\pm$0.01   \\
    GAE   & 0.79$\pm$0.01  & \textbf{0.98$\pm$0.00}  & 0.38$\pm$0.01  & 0.45$\pm$0.02  & 0.12$\pm$0.02  & 0.28$\pm$0.01   \\
    VGAE  & 0.80$\pm$0.01  & 0.97$\pm$0.00 & 0.39$\pm$0.01  & 0.44$\pm$0.02  & 0.18$\pm$0.03  & 0.28$\pm$0.00   \\
    MAE   & 0.82$\pm$0.01  & 0.52$\pm$0.00  & 0.30$\pm$0.02  & 0.24$\pm$0.05   & 0.03$\pm$0.00  & 0.22$\pm$0.01 \\
    scBERT & 0.77$\pm$0.01  & 0.47$\pm$0.00  & 0.39$\pm$0.01  & 0.37$\pm$0.03  & 0.26$\pm$0.01  & 0.28$\pm$0.00 \\
    MuSe-GNN & 0.75$\pm$0.01  & 0.78$\pm$0.02  & \textbf{0.53$\pm$0.01}  & \textbf{0.73$\pm$0.04}  & \textbf{0.65$\pm$0.01}  & \textbf{0.31$\pm$0.00}   \\
    \bottomrule
    \end{tabular}%
  
\end{table}%

\begin{table}[htbp]
  \centering
  \caption{Benchmark score table for Lung.}
  \label{Table: lung result}
    \begin{tabular}{lrrrrrrr}
    \toprule
    Methods & \multicolumn{1}{l}{ASW} & \multicolumn{1}{l}{AUC} & \multicolumn{1}{l}{iLISI} & \multicolumn{1}{l}{GC} & \multicolumn{1}{l}{CGR} & \multicolumn{1}{l}{NO} \\
    \midrule
    PCA   & 0.79$\pm$0.00  & 0.69$\pm$0.00  & 0.58$\pm$0.00  & 0.70$\pm$0.01  & 0.07$\pm$0.01  & 0.10$\pm$0.00  \\
    Gene2vec & 0.71$\pm$0.40  & 0.76$\pm$0.00  & 0.00$\pm$0.00  & \textbf{0.89$\pm$0.01}  & 0.00$\pm$0.00  & 0.01$\pm$0.00   \\
    GIANT & \textbf{0.90$\pm$0.01}  & 0.49$\pm$0.00  & 0.12$\pm$0.03  & 0.88$\pm$0.03  & 0.07$\pm$0.01  & 0.01$\pm$0.00   \\
    WSMAE & 0.76$\pm$0.01  & 0.69$\pm$0.01 & 0.45$\pm$0.03  & 0.77$\pm$0.03  & 0.05$\pm$0.01  & 0.09$\pm$0.01   \\
    GAE   & 0.73$\pm$0.03  & 0.86$\pm$0.00  & 0.23$\pm$0.04  & 0.78$\pm$0.03  & 0.02$\pm$0.00  & 0.07$\pm$0.01   \\
    VGAE  & 0.51$\pm$0.10  &\textbf{0.87$\pm$0.01} & 0.00$\pm$0.00  & 0.87$\pm$0.03  & 0.00$\pm$0.00  & 0.01$\pm$0.01  \\
    MAE   & 0.85$\pm$0.02  & 0.62$\pm$0.01  & 0.35$\pm$0.05  & 0.81$\pm$0.03  & 0.02$\pm$0.01  & 0.08$\pm$0.01 \\
    scBERT & 0.87$\pm$0.01  & 0.48$\pm$0.00  & 0.29$\pm$0.02  & 0.86$\pm$0.05  & 0.06$\pm$0.01  & 0.10$\pm$0.00 \\
    MuSe-GNN & 0.84$\pm$0.02  & 0.86$\pm$0.01  & \textbf{0.64$\pm$0.04}  & 0.87$\pm$0.03  & \textbf{0.37$\pm$0.03}  & \textbf{0.19$\pm$0.00}  \\
    \bottomrule
    \end{tabular}%
  
\end{table}%

\begin{table}[htbp]
  \centering
  \caption{Benchmark score table for Liver.}
  \label{Table: liver result}
    \begin{tabular}{lrrrrrrr}
    \toprule
    Methods & \multicolumn{1}{l}{ASW} & \multicolumn{1}{l}{AUC} & \multicolumn{1}{l}{iLISI} & \multicolumn{1}{l}{GC} & \multicolumn{1}{l}{CGR} & \multicolumn{1}{l}{NO} \\
    \midrule
    PCA   & 0.77$\pm$0.00  & 0.58$\pm$0.00  & 0.39$\pm$0.00  & 0.69$\pm$0.01  & 0.21$\pm$0.00  & 0.15$\pm$0.00  \\
    Gene2vec & 0.63$\pm$0.34  & 0.69$\pm$0.00  & 0.00$\pm$0.00  & 0.88$\pm$0.01  & 0.00$\pm$0.00  & 0.03$\pm$0.01  \\
    GIANT & \textbf{0.90$\pm$0.01}  & 0.55$\pm$0.00  & 0.40$\pm$0.02  & 0.44$\pm$0.03 & 0.11$\pm$0.01  & 0.00$\pm$0.00   \\
    WSMAE & 0.80$\pm$0.01  & 0.62$\pm$0.01  & 0.26$\pm$0.03  & 0.83$\pm$0.03  & 0.09$\pm$0.01  & 0.13$\pm$0.01  \\
    GAE   & 0.74$\pm$0.02  & 0.84$\pm$0.00  & 0.17$\pm$0.02  & 0.80$\pm$0.01  & 0.05$\pm$0.01  & 0.14$\pm$0.01   \\
    VGAE  & 0.49$\pm$0.05  & \textbf{0.87$\pm$0.00}  & 0.00$\pm$0.00  & 0.87$\pm$0.02  & 0.01$\pm$0.00  & 0.01$\pm$0.00   \\
    MAE   & 0.83$\pm$0.01  & 0.58$\pm$0.01  & 0.25$\pm$0.04  & 0.88$\pm$0.04 & 0.04$\pm$0.01  & 0.09$\pm$0.01 \\
    scBERT & 0.87$\pm$0.01  & 0.50$\pm$0.00  & 0.23$\pm$0.05  & \textbf{0.90$\pm$0.03}  & 0.11$\pm$0.02  & 0.14$\pm$0.00 \\
    MuSe-GNN & 0.86$\pm$0.01  & 0.78$\pm$0.03  & \textbf{0.65$\pm$0.02}  & 0.83$\pm$0.04  & \textbf{0.46$\pm$0.02}  & \textbf{0.18$\pm$0.00}  \\
    \bottomrule
    \end{tabular}%
  
\end{table}%

\begin{table}[htbp]
  \centering
  \caption{Benchmark score table for Kidney.}
  \label{Table: kidney result}
    \begin{tabular}{lrrrrrrr}
    \toprule
    Methods & \multicolumn{1}{l}{ASW} & \multicolumn{1}{l}{AUC} & \multicolumn{1}{l}{iLISI} & \multicolumn{1}{l}{GC} & \multicolumn{1}{l}{CGR} & \multicolumn{1}{l}{NO}  \\
    \midrule
    PCA   & 0.69$\pm$0.01  & 0.55$\pm$0.00  & 0.39$\pm$0.01  & 0.98$\pm$0.00  & 0.00$\pm$0.00  & 0.21$\pm$0.00 \\
    Gene2vec & 0.33$\pm$0.42  & 0.74$\pm$0.00  & 0.00$\pm$0.00  & 1.00$\pm$0.01  & 0.00$\pm$0.00  & 0.02$\pm$0.03   \\
    GIANT & \textbf{0.92$\pm$0.00}  & 0.55$\pm$0.00  & \textbf{0.89$\pm$0.01}  & 0.78$\pm$0.03  & 0.02$\pm$0.00  & 0.01$\pm$0.00  \\
    WSMAE & 0.74$\pm$0.05  & 0.68$\pm$0.01  & 0.08$\pm$0.04  & 0.99$\pm$0.00  & 0.01$\pm$0.00  & 0.15$\pm$0.04  \\
    GAE   & 0.52$\pm$0.10  & 0.86$\pm$0.00  & 0.01$\pm$0.01  & 0.98$\pm$0.01  & 0.00$\pm$0.00  & 0.05$\pm$0.02   \\
    VGAE  & 0.51$\pm$0.16  & \textbf{0.88$\pm$0.00}  & 0.00$\pm$0.00  & 0.98$\pm$0.01  & 0.00$\pm$0.00  & 0.00$\pm$0.01  \\
    MAE   & 0.84$\pm$0.04  & 0.52$\pm$0.01  & 0.17$\pm$0.07  & \textbf{1.00$\pm$0.00}  & 0.00$\pm$0.00  & 0.16$\pm$0.04 \\
    scBERT & 0.89$\pm$0.02  & 0.47$\pm$0.00  & 0.59$\pm$0.18  & \textbf{1.00$\pm$0.00}  & 0.05$\pm$0.01  & 0.30$\pm$0.00 \\
    MuSe-GNN & 0.88$\pm$0.03  & 0.71$\pm$0.02 & 0.84$\pm$0.04  & 0.98$\pm$0.02  & \textbf{0.31$\pm$0.04}  & \textbf{0.31$\pm$0.01} \\
    \bottomrule
    \end{tabular}%
  
\end{table}%

\begin{table}[htbp]
  \centering
  \caption{Benchmark score table for Thymus.}
  \label{Table: thymus result}
    \begin{tabular}{lrrrrrrr}
    \toprule
    Methods & \multicolumn{1}{l}{ASW} & \multicolumn{1}{l}{AUC} & \multicolumn{1}{l}{iLISI} & \multicolumn{1}{l}{GC} & \multicolumn{1}{l}{CGR} & \multicolumn{1}{l}{NO} \\
    \midrule
    PCA   & 0.76$\pm$0.01  & 0.79$\pm$0.00  & 0.93$\pm$0.00  & 0.44$\pm$0.05  & 0.00$\pm$0.00  & 0.00$\pm$0.00   \\
    Gene2vec & 0.00$\pm$0.00  & 0.75$\pm$0.00  & 0.00$\pm$0.00  & 0.58$\pm$0.03  & 0.00$\pm$0.00  & 0.00$\pm$0.00   \\
    GIANT & \textbf{0.88$\pm$0.02}  & 0.55$\pm$0.00  & \textbf{0.98$\pm$0.01}  & \textbf{0.80$\pm$0.02}  & 0.00$\pm$0.00  & 0.00$\pm$0.00   \\
    WSMAE & 0.81$\pm$0.03  & 0.74$\pm$0.01  & 0.93$\pm$0.06  & 0.48$\pm$0.06  & 0.00$\pm$0.00  & 0.00$\pm$0.00   \\
    GAE   & 0.72$\pm$0.12  & 0.81$\pm$0.00  & 0.89$\pm$0.10  & 0.69$\pm$0.05  & 0.00$\pm$0.00  & 0.00$\pm$0.00   \\
    VGAE  & 0.77$\pm$0.05  & \textbf{0.84$\pm$0.01}  & 0.89$\pm$0.05  & 0.44$\pm$0.10  & 0.00$\pm$0.00  & 0.00$\pm$0.00 \\
    MAE   & 0.77$\pm$0.28  & 0.59$\pm$0.01  & 0.94$\pm$0.05  & 0.33$\pm$0.05   & 0.00$\pm$0.00  & 0.00$\pm$0.00 \\
    scBERT & 0.09$\pm$0.27  & 0.50$\pm$0.00  & 0.41$\pm$0.18  & 0.42$\pm$0.04  & 0.00$\pm$0.00  & 0.00$\pm$0.00 \\
    MuSe-GNN & 0.71$\pm$0.05  & 0.80$\pm$0.02  & \textbf{0.98$\pm$0.01}  & 0.65$\pm$0.07  & \textbf{0.01$\pm$0.00}  & 0.00$\pm$0.00  \\
    \bottomrule
    \end{tabular}%
  
\end{table}%

\begin{table}[htbp]
  \centering
  \caption{Benchmark score table for Spleen.}
  \label{Table: spleen result}
    \begin{tabular}{lrrrrrrr}
    \toprule
    Methods & \multicolumn{1}{l}{ASW} & \multicolumn{1}{l}{AUC} & \multicolumn{1}{l}{iLISI} & \multicolumn{1}{l}{GC} & \multicolumn{1}{l}{CGR} & \multicolumn{1}{l}{NO} \\
    \midrule
    PCA   & 0.81$\pm$0.01  & 0.56$\pm$0.00  & 0.47$\pm$0.00  & 0.73$\pm$0.01  & 0.23$\pm$0.01  & 0.17$\pm$0.00   \\
    Gene2vec & 0.69$\pm$0.36  & 0.68$\pm$0.01  & 0.00$\pm$0.00  & 0.85$\pm$0.01  & 0.00$\pm$0.00  & 0.01$\pm$0.01  \\
    GIANT & \textbf{0.88$\pm$0.01}  & 0.55$\pm$0.00  & 0.42$\pm$0.02  & 0.45$\pm$0.02  & 0.10$\pm$0.00  & 0.01$\pm$0.00   \\
    WSMAE & 0.79$\pm$0.01  & 0.61$\pm$0.01  & 0.26$\pm$0.03  & 0.86$\pm$0.02  & 0.08$\pm$0.01  & 0.15$\pm$0.01   \\
    GAE   & 0.75$\pm$0.02  & 0.84$\pm$0.00  & 0.18$\pm$0.02  & 0.81$\pm$0.01  & 0.05$\pm$0.01  & 0.16$\pm$0.01   \\
    VGAE  & 0.52$\pm$0.05  & \textbf{0.86$\pm$0.00}  & 0.00$\pm$0.00  & 0.85$\pm$0.03  & 0.00$\pm$0.00  & 0.01$\pm$0.00 \\
    MAE   & 0.84$\pm$0.02  & 0.56$\pm$0.01  & 0.30$\pm$0.04  & 0.92$\pm$0.02  & 0.03$\pm$0.00  & 0.12$\pm$0.00 \\
    scBERT & 0.87$\pm$0.01  & 0.50$\pm$0.00  & 0.34$\pm$0.06  & 0.94$\pm$0.02 & 0.11$\pm$0.01  & 0.16$\pm$0.00 \\
    MuSe-GNN & 0.86$\pm$0.01  & 0.79$\pm$0.02  & \textbf{0.70$\pm$0.02}  & \textbf{0.89$\pm$0.03}  & \textbf{0.47$\pm$0.01}  & \textbf{0.19$\pm$0.00}   \\
    \bottomrule
    \end{tabular}%
  
\end{table}%

\begin{table}[htbp]
  \centering
  \caption{Benchmark score table for Pancreas.}
  \label{Table: pancreas result}
    \begin{tabular}{lrrrrrrr}
    \toprule
    Methods & \multicolumn{1}{l}{ASW} & \multicolumn{1}{l}{AUC} & \multicolumn{1}{l}{iLISI} & \multicolumn{1}{l}{GC} & \multicolumn{1}{l}{CGR} & \multicolumn{1}{l}{NO}  \\
    \midrule
    PCA   & 0.61$\pm$0.01  & 0.62$\pm$0.00  & 0.64$\pm$0.00  & 0.57$\pm$0.01  & 0.00$\pm$0.00  & 0.01$\pm$0.00   \\
    Gene2vec & 0.17$\pm$0.36  & 0.73$\pm$0.00  & 0.00$\pm$0.00  & 0.51$\pm$0.05  & 0.00$\pm$0.00  & 0.00$\pm$0.00   \\
    GIANT & \textbf{0.91$\pm$0.01}  & 0.55$\pm$0.00  & \textbf{0.97$\pm$0.03}  & 0.65$\pm$0.05  & 0.00$\pm$0.00  & 0.00$\pm$0.00   \\
    WSMAE & 0.67$\pm$0.09  & 0.70$\pm$0.01  & 0.66$\pm$0.10  & 0.49$\pm$0.06  & 0.00$\pm$0.00  & 0.00$\pm$0.00   \\
    GAE   & 0.09$\pm$0.27  & 0.82$\pm$0.00  & 0.90$\pm$0.07  & \textbf{0.65$\pm$0.03}  & 0.00$\pm$0.00  & 0.00$\pm$0.00   \\
    VGAE  & 0.37$\pm$0.40  & \textbf{0.83$\pm$0.01}  & 0.13$\pm$0.10  & 0.51$\pm$0.04  & 0.00$\pm$0.00  & 0.00$\pm$0.00   \\
    MAE   & 0.71$\pm$0.11  & 0.56$\pm$0.01  & 0.83$\pm$0.07  & 0.37$\pm$0.05   & 0.00$\pm$0.00  & 0.00$\pm$0.00 \\
    scBERT & 0.76$\pm$0.05  & 0.47$\pm$0.00  & 0.54$\pm$0.15  & 0.39$\pm$0.04  & 0.00$\pm$0.00  & \textbf{0.01$\pm$0.00} \\
    MuSe-GNN & 0.72$\pm$0.08  & 0.65$\pm$0.01  & 0.95$\pm$0.13  & 0.62$\pm$0.06  & \textbf{0.01$\pm$0.00}  & 0.00$\pm$0.00   \\
    \bottomrule
    \end{tabular}%
  
\end{table}%

\begin{table}[htbp]
  \centering
  \caption{Benchmark score table for Cerebrum.}
  \label{Table: cerebrum result}
    \begin{tabular}{lrrrrrrr}
    \toprule
    Methods & \multicolumn{1}{l}{ASW} & \multicolumn{1}{l}{AUC} & \multicolumn{1}{l}{iLISI} & \multicolumn{1}{l}{GC} & \multicolumn{1}{l}{CGR} & \multicolumn{1}{l}{NO} \\
    \midrule
    PCA   & 0.82$\pm$0.01  & 0.63$\pm$0.00  & 0.34$\pm$0.00  & 0.95$\pm$0.00  & 0.12$\pm$0.00  & 0.18$\pm$0.00   \\
    Gene2vec & 0.07$\pm$0.23  & 0.69$\pm$0.00  & 0.00$\pm$0.00  & \textbf{0.99$\pm$0.00}  & 0.00$\pm$0.00  & 0.01$\pm$0.01   \\
    GIANT & 0.89$\pm$0.00  & 0.55$\pm$0.00  & 0.40$\pm$0.02  & 0.51$\pm$0.02  & 0.03$\pm$0.00  & 0.01$\pm$0.00   \\
    WSMAE & 0.80$\pm$0.02  & 0.67$\pm$0.01  & 0.21$\pm$0.04  & 0.97$\pm$0.01  & 0.03$\pm$0.01  & 0.13$\pm$0.02   \\
    GAE   & 0.68$\pm$0.04  & 0.83$\pm$0.00  & 0.03$\pm$0.01  & 0.95$\pm$0.01  & 0.02$\pm$0.00  & 0.13$\pm$0.01   \\
    VGAE  & 0.47$\pm$0.09  & \textbf{0.84$\pm$0.01} & 0.00$\pm$0.00  & 0.98$\pm$0.01  & 0.00$\pm$0.00  & 0.01$\pm$0.00   \\
MAE   & 0.83$\pm$0.02  & 0.53$\pm$0.01 & 0.20$\pm$0.05  & 0.99$\pm$0.01    & 0.02$\pm$0.01  & 0.14$\pm$0.01 \\
    scBERT & 0.89$\pm$0.01  & 0.48$\pm$0.00 & 0.50$\pm$0.07  & \textbf{0.99$\pm$0.00}  & 0.10$\pm$0.01  & 0.17$\pm$0.00 \\
    MuSe-GNN & \textbf{0.90$\pm$0.01}  & 0.73$\pm$0.02  & \textbf{0.79$\pm$0.03}  & 0.99$\pm$0.01  & \textbf{0.54$\pm$0.01}  & \textbf{0.21$\pm$0.00}  \\
    \bottomrule
    \end{tabular}%
  
\end{table}%

\begin{table}[htbp]
  \centering
  \caption{Benchmark score table for Cerebellum.}
  \label{Table: cerebellum result}
    \begin{tabular}{lrrrrrrr}
    \toprule
    Methods & \multicolumn{1}{l}{ASW} & \multicolumn{1}{l}{AUC} & \multicolumn{1}{l}{iLISI} & \multicolumn{1}{l}{GC} & \multicolumn{1}{l}{CGR} & \multicolumn{1}{l}{NO}  \\
    \midrule
    PCA   & 0.82$\pm$0.00  & 0.63$\pm$0.00  & 0.34$\pm$0.00  & 0.95$\pm$0.00  & 0.12$\pm$0.01  & 0.18$\pm$0.00   \\
    Gene2vec & 0.31$\pm$0.40  & 0.71$\pm$0.00  & 0.00$\pm$0.00  & \textbf{1.00$\pm$0.00}  & 0.00$\pm$0.00  & 0.02$\pm$0.02   \\
    GIANT & 0.89$\pm$0.00  & 0.55$\pm$0.00  & 0.34$\pm$0.03  & 0.73$\pm$0.02  & 0.03$\pm$0.00  & 0.01$\pm$0.00   \\
    WSMAE & 0.83$\pm$0.02  & 0.66$\pm$0.01  & 0.26$\pm$0.04  & \textbf{1.00$\pm$0.00}  & 0.06$\pm$0.01  & 0.29$\pm$0.02   \\
    GAE   & 0.74$\pm$0.02  & \textbf{0.83$\pm$0.00}  & 0.11$\pm$0.08  & 0.99$\pm$0.01  & 0.07$\pm$0.04  & 0.25$\pm$0.04   \\
    VGAE  & 0.19$\pm$0.26  & 0.83$\pm$0.01  & 0.00$\pm$0.00  & 0.99$\pm$0.01  & 0.00$\pm$0.00  & 0.00$\pm$0.00   \\
    MAE   & 0.86$\pm$0.01  & 0.51$\pm$0.00  & 0.16$\pm$0.03  & \textbf{1.00$\pm$0.00}  & 0.02$\pm$0.00  & 0.26$\pm$0.03 \\
    scBERT & 0.90$\pm$0.01  & 0.47$\pm$0.00  & 0.54$\pm$0.06  & \textbf{1.00$\pm$0.00}  & 0.15$\pm$0.02  & 0.31$\pm$0.01 \\
    MuSe-GNN & \textbf{0.92$\pm$0.01}  &0.64$\pm$0.02  & \textbf{0.86$\pm$0.02} & 0.98$\pm$0.02  & \textbf{0.65$\pm$0.00}  & \textbf{0.38$\pm$0.00}   \\
    \bottomrule
    \end{tabular}%
  
\end{table}%

\begin{table}[htbp]
  \centering
  \caption{Benchmark score table for PBMC.}
  \label{Table: pbmc result}
    \begin{tabular}{lrrrrrr}
    \toprule
    Methods & \multicolumn{1}{l}{ASW} & \multicolumn{1}{l}{AUC} & \multicolumn{1}{l}{iLISI} & \multicolumn{1}{l}{GC} & \multicolumn{1}{l}{CGR} & \multicolumn{1}{l}{NO} \\
    \midrule
    PCA   & 0.86$\pm$0.01 & 0.53$\pm$0.00  & 0.46$\pm$0.00  & 0.66$\pm$0.01  & 0.22$\pm$0.01  & 0.13$\pm$0.00 \\
    GENE2VEC & 0.00$\pm$0.00  & 0.59$\pm$0.00  & 0.00$\pm$0.00  & 0.92$\pm$0.01  & 0.00$\pm$0.00  & 0.00$\pm$0.00 \\
    GIANT & 0.85$\pm$0.01  & 0.55$\pm$0.00  & 0.38$\pm$0.02  & 0.55$\pm$0.03  & 0.04$\pm$0.00  & 0.00$\pm$0.00 \\
    WSMAE & 0.81$\pm$0.01  & 0.58$\pm$0.01  & 0.29$\pm$0.02  & 0.81$\pm$0.05  & 0.10$\pm$0.02  & 0.12$\pm$0.01 \\
    GAE   & 0.74$\pm$0.01  & 0.82$\pm$0.00  & 0.15$\pm$0.03  & 0.79$\pm$0.03  & 0.06$\pm$0.02  & 0.12$\pm$0.01 \\
    VGAE  & 0.67$\pm$0.04  & \textbf{0.89$\pm$0.00}  & 0.00$\pm$0.00  & 0.93$\pm$0.02  & 0.02$\pm$0.00  & 0.01$\pm$0.00 \\
    MAE   & 0.85$\pm$0.02  & 0.53$\pm$0.01  & 0.30$\pm$0.05  & \textbf{0.94$\pm$0.03}  & 0.02$\pm$0.00  & 0.10$\pm$0.01 \\
    scBERT & \textbf{0.88$\pm$0.01}  & 0.52$\pm$0.00  & 0.48$\pm$0.05  & 0.93$\pm$0.03  & 0.15$\pm$0.02  & 0.11$\pm$0.00 \\
    MuSe-GNN & 0.86$\pm$0.01  & 0.79$\pm$0.02  & \textbf{0.75$\pm$0.02}  & 0.88$\pm$0.02  & \textbf{0.54$\pm$0.01}  & \textbf{0.18$\pm$0.00} \\
    \bottomrule
    \end{tabular}%
\end{table}%

\subsection{Details about function cluster identification}
\label{appexp: function cluster identi}
In this paper, we utilized Leiden cluster method \cite{traag2019louvain} to identify genes with common functions based on the value of their embeddings. The genes with distance smaller than a community-discovering based threshold (known as co-embedded genes) will be identified into a group. Here we sample 100 genes in our final gene embedding result from heart tissue to illustrate the distribution of such distance.
\begin{figure}[ht]
    \centering
    \includegraphics[width=0.8\textwidth]{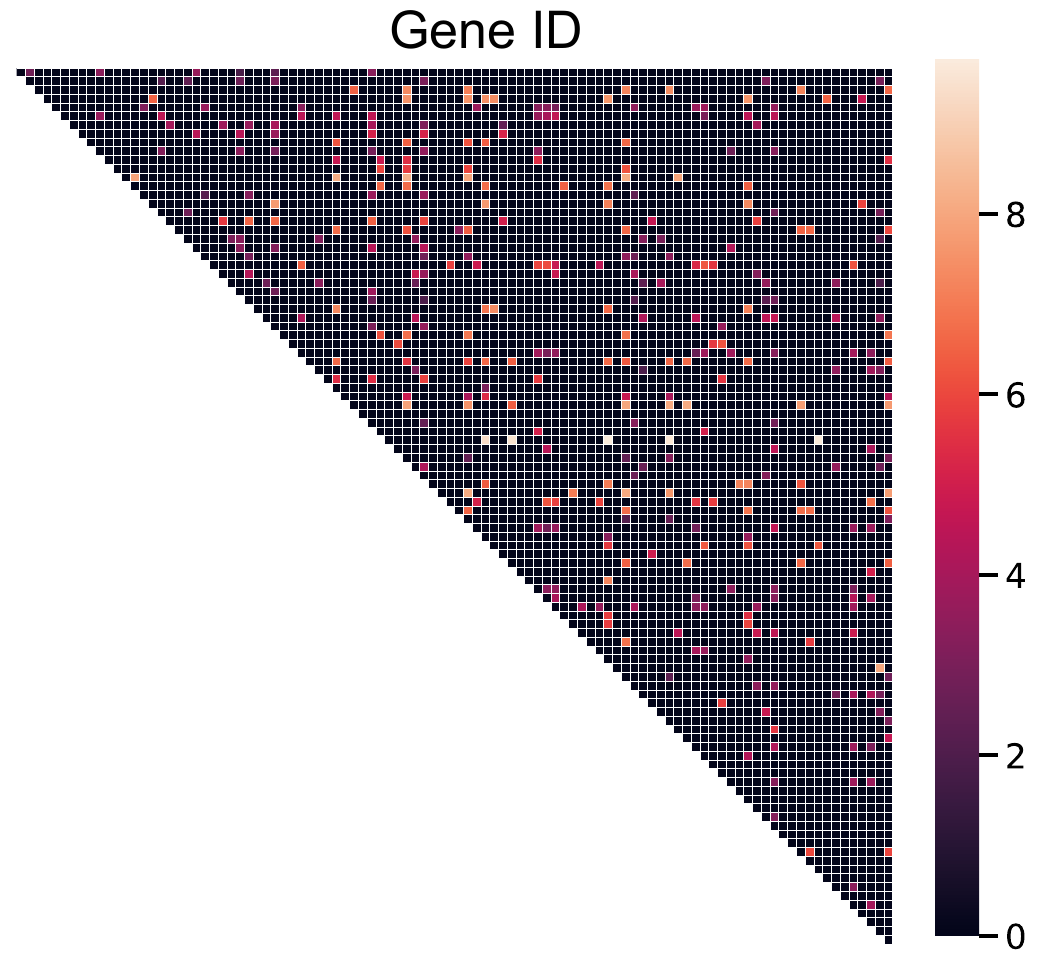}
    \caption{Heatmap of gene embedding distances. The heatmap color represents the distance between genes, with brighter colors indicating greater distances and more distinct functions.
    \label{fig:heart heatmap}}
\end{figure}
Based on Figure \ref{fig:heart heatmap}, we can conclude that our method has the ability to identify genes with similar embeddings, and further, it can be used to derive genes with similar functions.

All of the experiments of this paper are finished based on Intel Xeon Gold 6240 Processor and one NVIDIA A100 GPU. The RAM setting is 30 GB.

\section{Extra Discussion about Model Design}
\label{Extra model discussion}
This section aims to demonstrate the reasoning behind our model design in greater details. As a novel graph neural network model, MuSe-GNN offers three significant advantages.

Firstly, our method's graph construction process is robust and reliable. Graph construction is a critical step for GNN models \cite{zhu2021graph}, and we place considerable emphasis on both the graph construction method and data quality throughout our work. In single-cell data analysis research, some researchers model cell-cell similarity based on the nearest neighbor graph \cite{wang2021scgnn, van2018recovering}. This approach may be affected by batch effects resulting from different sequencing profiles, leading to incorrect conclusions. For instance, in scRNA-seq data imputation tasks, such a design may result in a high false positive rate \cite{andrews2018false}. A gene co-expression network derived from a robust statistical framework is more trustworthy, and biological experiments can validate the co-expression performance of various genes. Moreover, to understand the difference of co-expression networks from different tissues or omics data, we calculate the Edge Signal-Noise Ratio (ESNR) \cite{dong2023towards} for each network, and figure out that the co-expression network based on scRNA-seq has the best quality, while the co-expression networks from other omics have very small ESNR. Such conclusions are also consistent with the model performance. 

Secondly, our model utilizes a sophisticated and stable training strategy. By employing the cosine similarity penalty, we ensure that genes with the same functions from different tissues or techniques can be co-embedded in similar positions. Additionally, we use graph auto-encoder and contrastive learning strategies to retain the original biological information of various datasets. Each of our training strategies serves a specific purpose.

Finally, our method showcases the high effectiveness of MMML in addressing heterogeneous data learning challenges. MMML can successfully learn similar information across multimodal data and represent it in a unified space through model fusion \cite{liang2022foundations}. This approach can be applied to more complex data studies in the future, such as integrating data from Magnetic Resonance Imaging (MRI) results, spatial resolution image information, and current expression profiles into a single space. Furthermore, GNN remains the primary model for handling tasks related to graph-structured data. By combining these methods and uncovering biological insights, our research contributes to a deeper understanding of efficient human genome representation.

\section{Selecting anchors of common functional genes}
\label{appendix: commmon genes}

In this paper, we employ explicit common functional genes as anchors across different datasets to implement regularization in our training process. The common functional genes we select can potentially recover a portion of the functional overlap of the genes, while the remaining implicit genes can be recovered during the training process.

\begin{figure}[ht]
    \centering
    \includegraphics[width=1\textwidth]{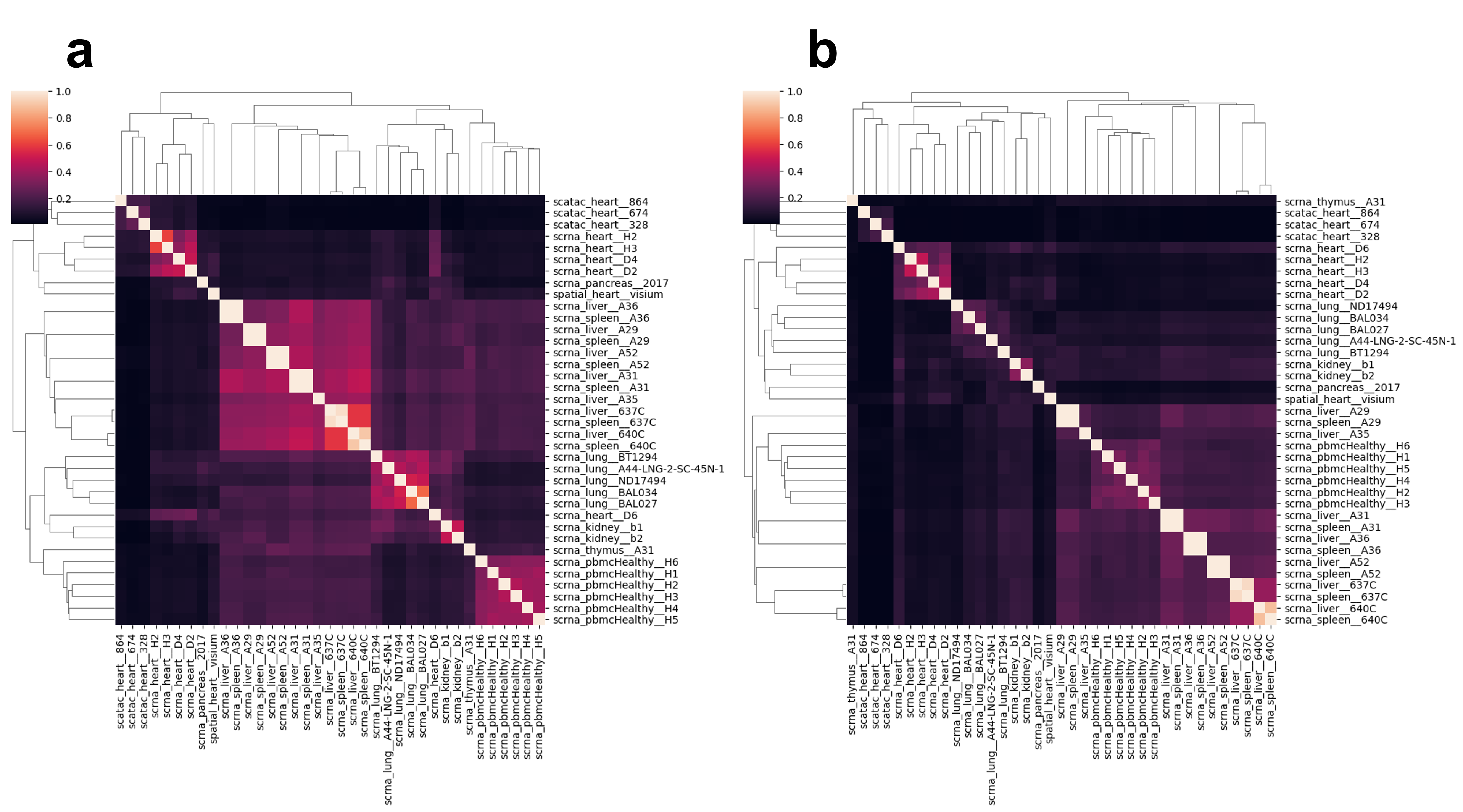}
    \caption{Cluster heatmaps of anchor genes for multimodal data. \textbf{(a)} represents the cluster heatmap based on HVGs overlap, \textbf{(b)} represents the cluster heatmap based on the neighbor genes overlap of HVGs.
    \label{fig:sup similar 2}}
\end{figure}

\begin{figure*}[ht]
    \centering
    \includegraphics[width=1\textwidth]{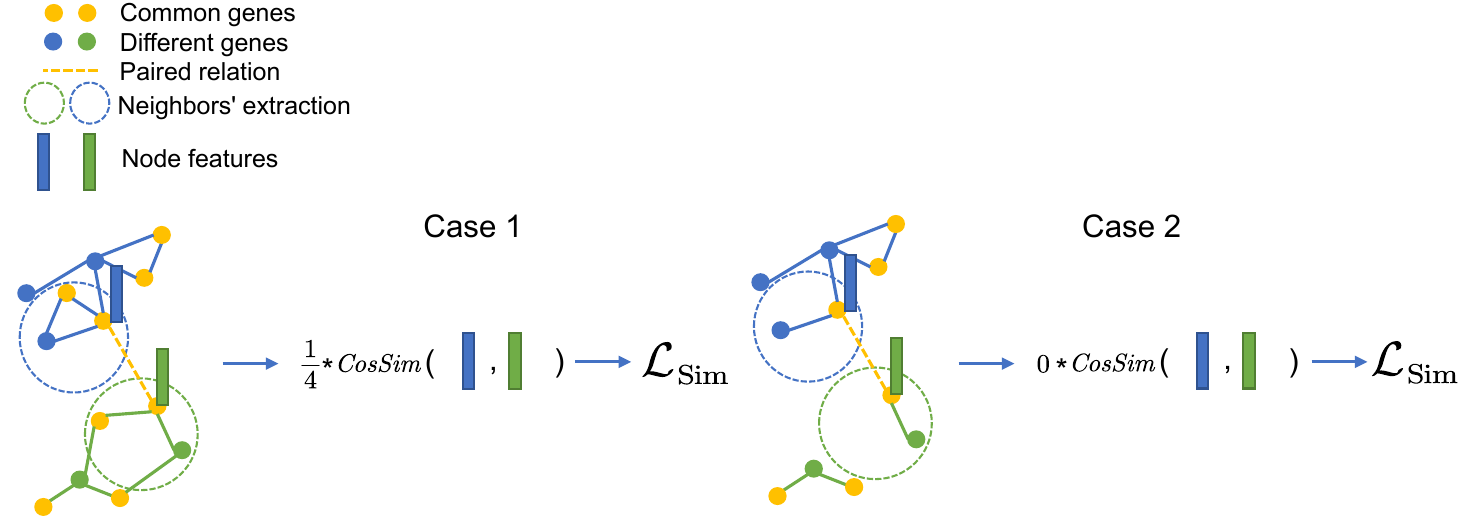}
    \caption{The need for weight similarity learning illustrated by two cases.
    \label{fig:illu sim learning}}
\end{figure*}

\begin{figure*}[ht]
    \centering
    \includegraphics[width=1\textwidth]{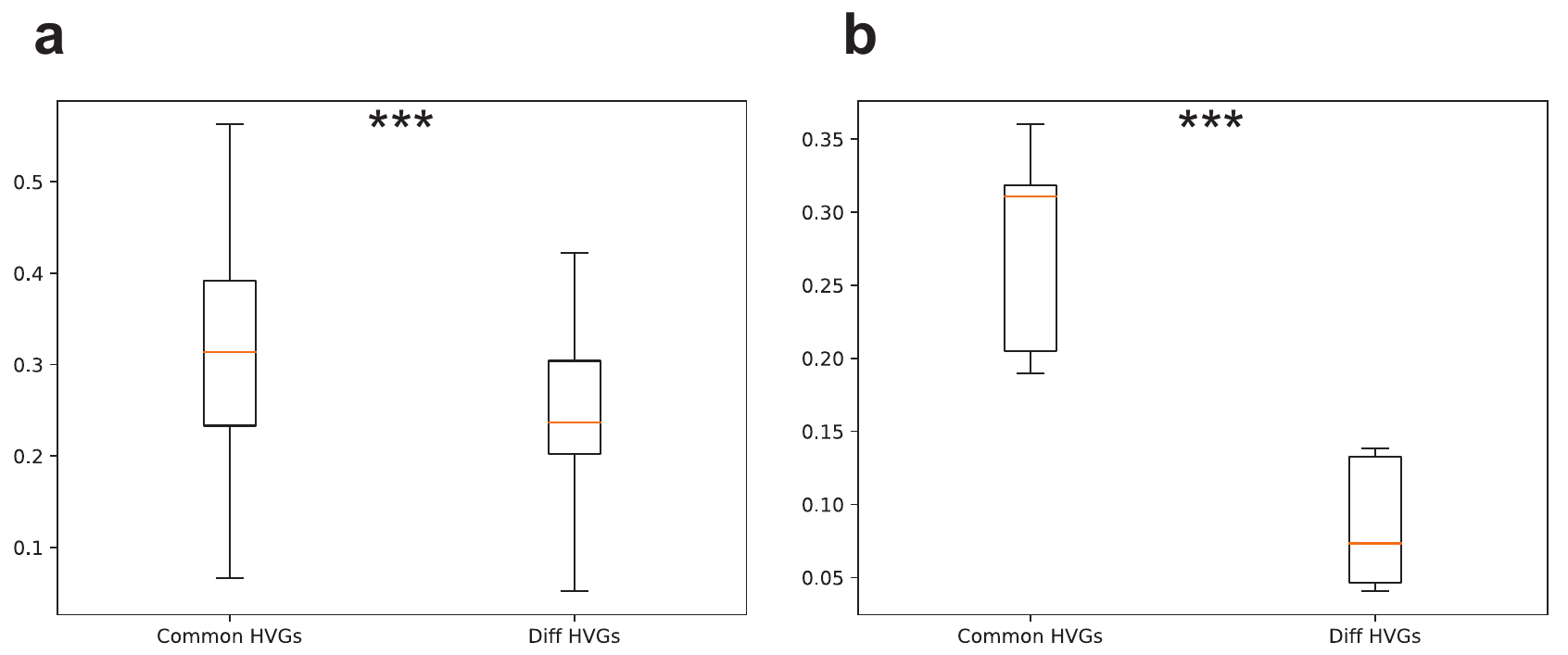}
    \caption{Boxplots for the neighbor overlap score of HVGs from different tissues. We used Kolmogorov–Smirnov test \cite{virtanen2020scipy} to compare the difference between the two groups of genes. The significance level of group difference is shown by the stars ($***: \text{P-value} \leq 0.01$). \textbf{(a)} represent the distribution of overlap scores for heart tissue, grouped by common HVGs and different HVGs. \textbf{(b)} represent the distribution of overlap score for cerebrum tissue, grouped by common HVGs and different HVGs.
    \label{fig:boxplot shown}}
\end{figure*}

In this paper, we choose to use common highly variable genes (HVGs) combined from different datasets as anchors, and employ their co-expression network overlap as weights. An example is illustrated in Figure \ref{fig:illu sim learning} case 1. In addition to the proof provided by the ablation test results, there are four reasons for this choice: 1. HVGs represent the active genes of a expression profile (dataset), which are correlated to gene functions. 2. For datasets from the same tissue, the overlap of HVGs and the overlap of neighbor genes (known as the co-expressed genes in the same graph) of HVGs pair are similar. This demonstrates a type of consistency in biological function inference. 3. For datasets from different modalities (tissues, techniques, etc.), only using the overlap of HVGs is not reliable. The overlap of neighbor genes of HVGs pair from different tissues shows functional similarity based on hierarchical clustering. We can observe the clear difference of HVGs overlap score and neighbors overlap score of HVGs by comparing Figure \ref{fig:sup similar 2} (a) to (b). Furthermore, we can consider an extreme case, illustrated in Figure \ref{fig:illu sim learning} case 2. If two HVGs from different tissues do not contain shared neighbors, the training process will be processed for such genes without weighted constraints. 4. Since it is quite challenging to select all the true genes with similar functions, we can perform a trade-off between learning the similarity of potential functionally similar genes and learning the difference of potential functionally distinct genes. The neighbor genes overlap in common HVGs is significantly different from the overlap in different HVGs for graphs from the same tissue, as shown in Figure \ref{fig:boxplot shown}. Such statistical results also support our choice of using common HVGs as anchors.

In conclusion, we aim to generate similar embeddings for functionally related genes, incorporating common HVGs and their neighbors in co-expression networks, by utilizing the regularization term associated with common HVGs. Conversely, embeddings for functionally distinct genes should differ, enabling their identification through community detection-based clustering algorithms.

\section{Metrics details}
\label{appendix: metrics details}
Here we assume that HVGs from the same tissues with the same name have similar functions across different datasets, and we also consider evaluating the preservation of the known gene co-expression relation by using node similarity. These metrics include:
\begin{itemize}
    \item edge AUC: Edge Area under the ROC Curve (AUC) is a metric widely used in the edge prediction task. We calculate the AUC score between the multiplication of normalized embeddings with Sigmoid threshold ($Sigmoid(zz^T)$) and the true edge relation. This metric can reflect the co-expression information we have in our embedding space.

    \item common genes ASW: Common genes ASW represents the ASW score we calculated based on common genes across different tissues. Since the overlap genes of different datasets are not very large, based on the assumption that different highly variable genes have similar functions in different datasets, we can utilize common genes ASW as a metric to evaluate our integration performance.

    \item common genes Graph connectivity: Graph connectivity is a metric to evaluate the connectivity based on the ratio between the number of nodes in the largest connected component of the graph constructed by the gene cluster and the number of nodes belonging to the graph. We calculate the graph connectivity score for each gene cluster and take the average.

    \item common genes iLISI: The inverse Simpson’s index (iLISI) determines the number of common genes that can be drawn from a neighbor list before one dataset (or datasets from the same tissue) is observed twice. Therefore, this metric can be used to evaluate the integration level of the common genes across different datasets or tissues based on the choice of our index type.

    \item common genes ratio: The common genes ratio is defined based on the weighted average of one minus the ratio of unique genes from one cluster to the total genes in the cluster across all the clusters. We utilize the Leiden method to generate the clusters. This metric can be used to evaluate the level of integration of the same functional genes by a given method. 

    \item neighbors overlap: The neighbor overlap value is defined based on the neighbors of one gene in the co-expression network. For each cluster, we calculate the similarity of genes' neighbors among different datasets and perform a weighted average of the similarity. A larger value means that the method is able to integrate similar genes.
\end{itemize}

\section{Shared transcription factors and pathways analysis}
\label{appendix: tf analysis}
\begin{figure}[H]
    \centering
    \includegraphics[width=1\textwidth]{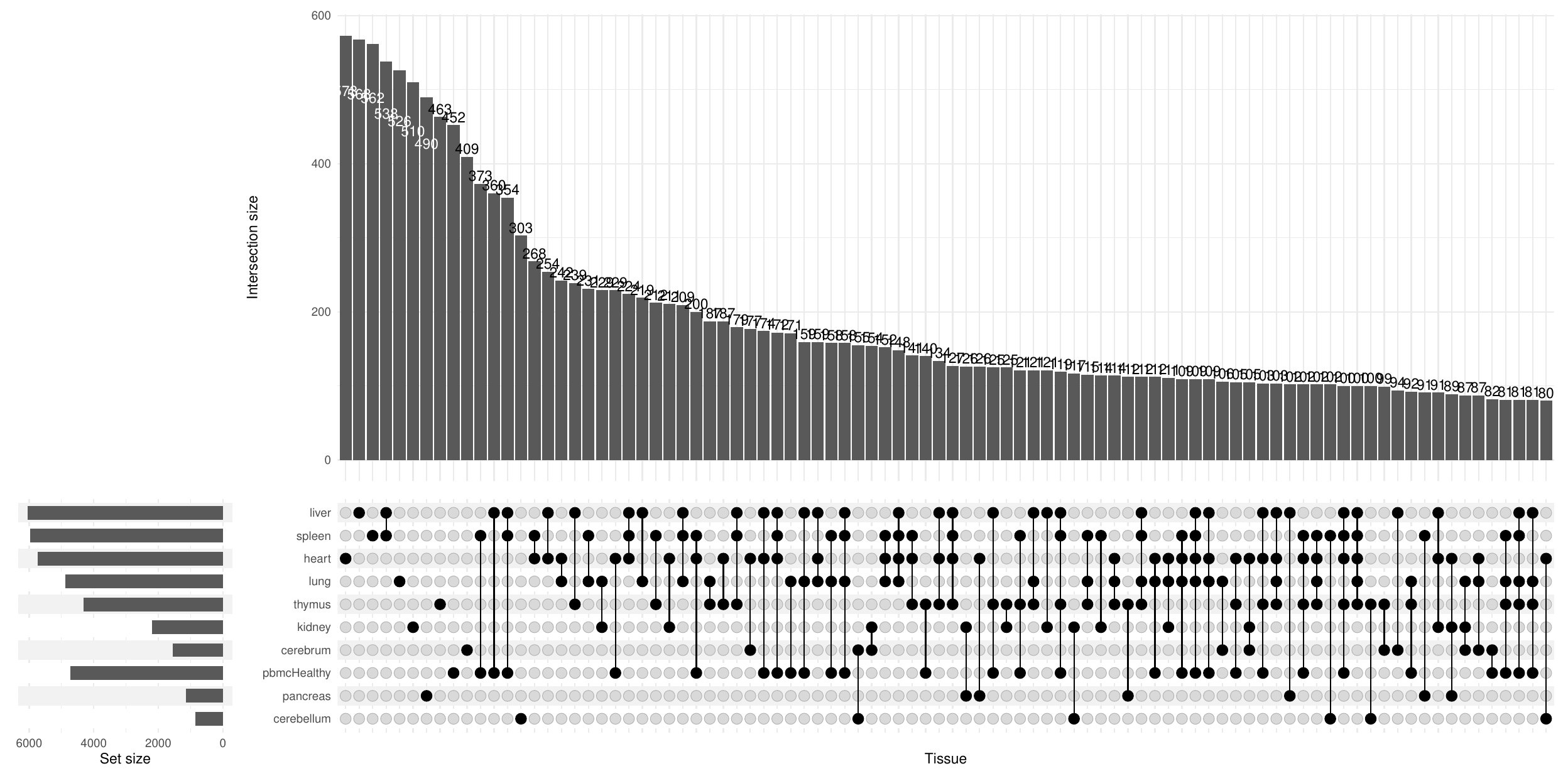}
    \caption{Results of common TFs discovery.
    \label{fig:common factor}}
\end{figure}

Transcription factors (TFs) are crucial in transcriptional regulation \cite{han2018trrust}. By examining the overlap of TFs across diverse datasets, we can assess the similarities between distinct modal data in terms of transcriptional regulation. We employed the TRRUST v2 database \cite{han2018trrust} to identify regulators for each gene within the gene embedding space. Subsequently, for every cluster, we extracted the transcription factors associated with the genes and computed the transcription factor overlap for the various multimodal data linked to the genes. We iteratively repeated this process for each cluster and documented the number of shared transcription factors based on different combinations of biological multimodal data. The results are illustrated in Figure \ref{fig:common factor}.
Figure \ref{fig:common factor} reveals that tissues with similar functions exhibit greater TF overlap, such as the heart and lung group, and the spleen and liver group. These findings demonstrate that our generated embeddings effectively capture the similarities between various data types. ComplexUpset \cite{krassowski2020complexupset} was utilized to create this figure.

\begin{figure}[H]
    \centering
    \includegraphics[width=1\textwidth]{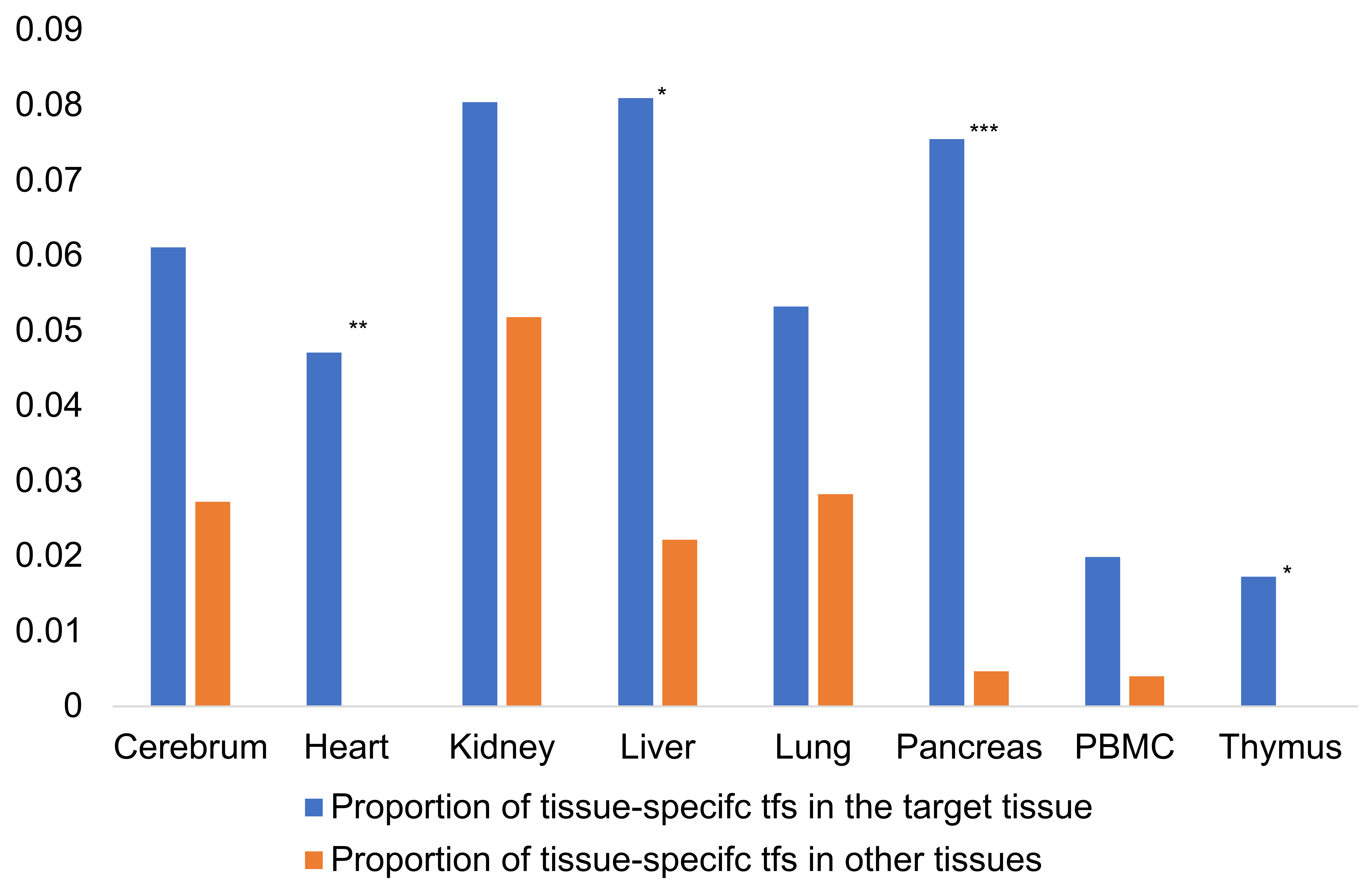}
    \caption{The proportion of tissue-specific TFs enrichment in the target tissue (blue) and in other tissues (orange). Here the significant enrichment in blue groups are marked by stars ($*: \text{P-value} \leq 0.1; **: \text{P-value} \leq 0.05; ***: \text{P-value} \leq 0.01;$).
    \label{fig:tf by tissue}}
\end{figure}
Additionally, we can emphasize the enrichment status of tissue-specific TFs, as depicted in Figure \ref{fig:tf by tissue}. We utilized the TF information from \cite{xu2022tf} to compute the enrichment proportion for tissue-specific TFs enriched in the given tissue and other tissues. A one-tailed Fisher's exact test was also conducted to assess the significance of enrichment differences. The figure suggests that our gene embeddings consistently enrich tissue-specific TFs across all evaluated tissues. Notably, for some tissues, the enrichment is significant, which aligns with the conclusions from \cite{chen2022unified}. However, in other tissues, the difference is not significant. A potential explanation is that such tissues may play a critical role in the interaction between different tissues for specific biological processes.

\begin{figure}[H]
    \centering
    \includegraphics[width=1\textwidth]{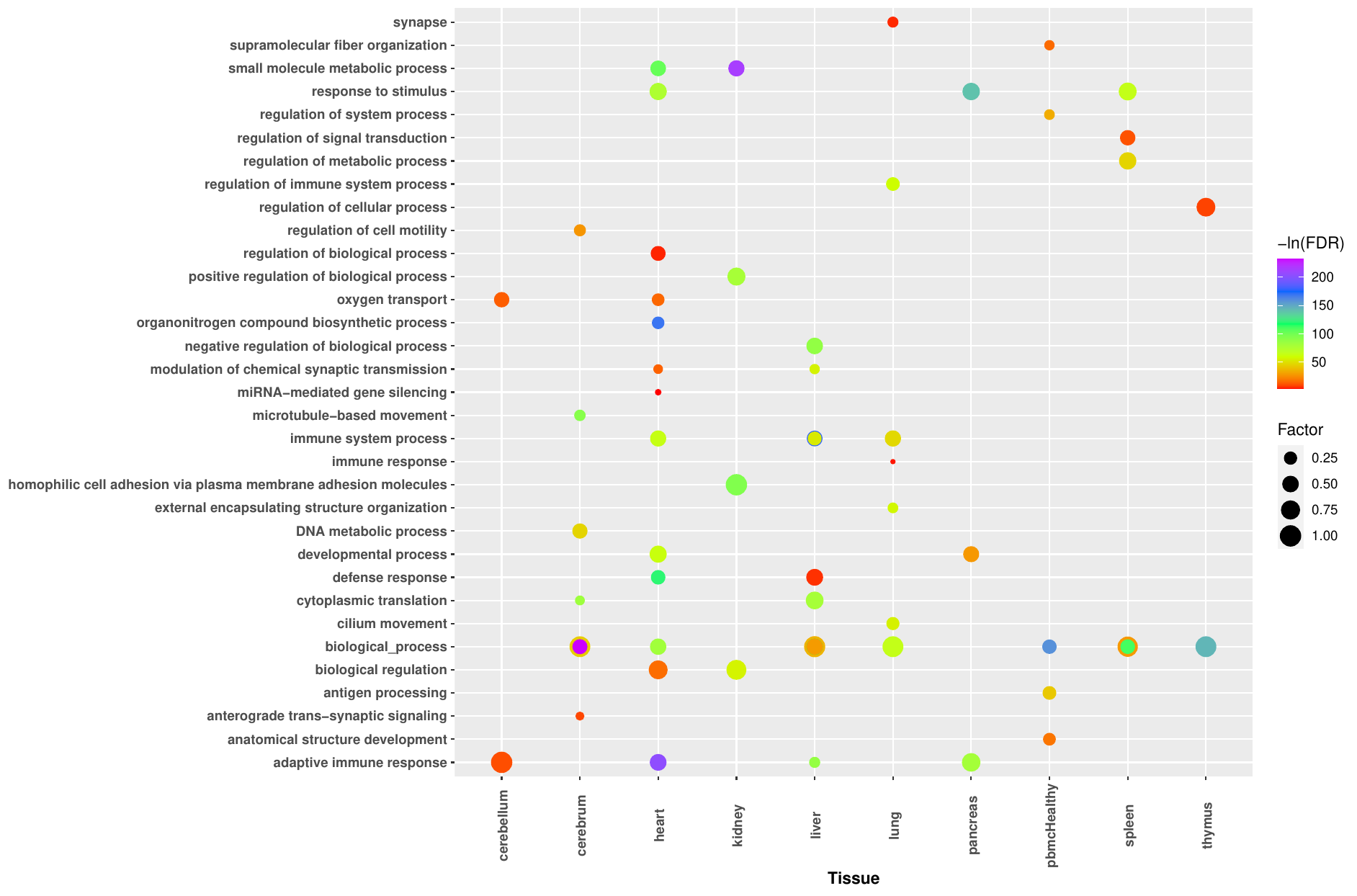}
    \caption{Top1 enriched pathways across different gene function clusters grouped by major tissue in each cluster. Here the color of bubbles represents the $-ln(FDR)$ value and the size of bubbles represents the value of rich factor for different pathways.
    \label{fig:all pathways}}
\end{figure}

In Figure \ref{fig:all pathways}, we determined the most enriched pathway for each cluster and classified it into the most representative tissue within that cluster by GOEA. The representative tissue was chosen based on the highest proportion of the source tissue for genes. This figure allows us to conclude that the heart tissue is the most versatile in the human body, followed by liver and lung tissues, which are also crucial tissues in humans.

\section{Multi-species gene embeddings}
\label{appendix: multispecies}
\begin{figure}[ht]
    \centering
    \includegraphics[width=1\textwidth]{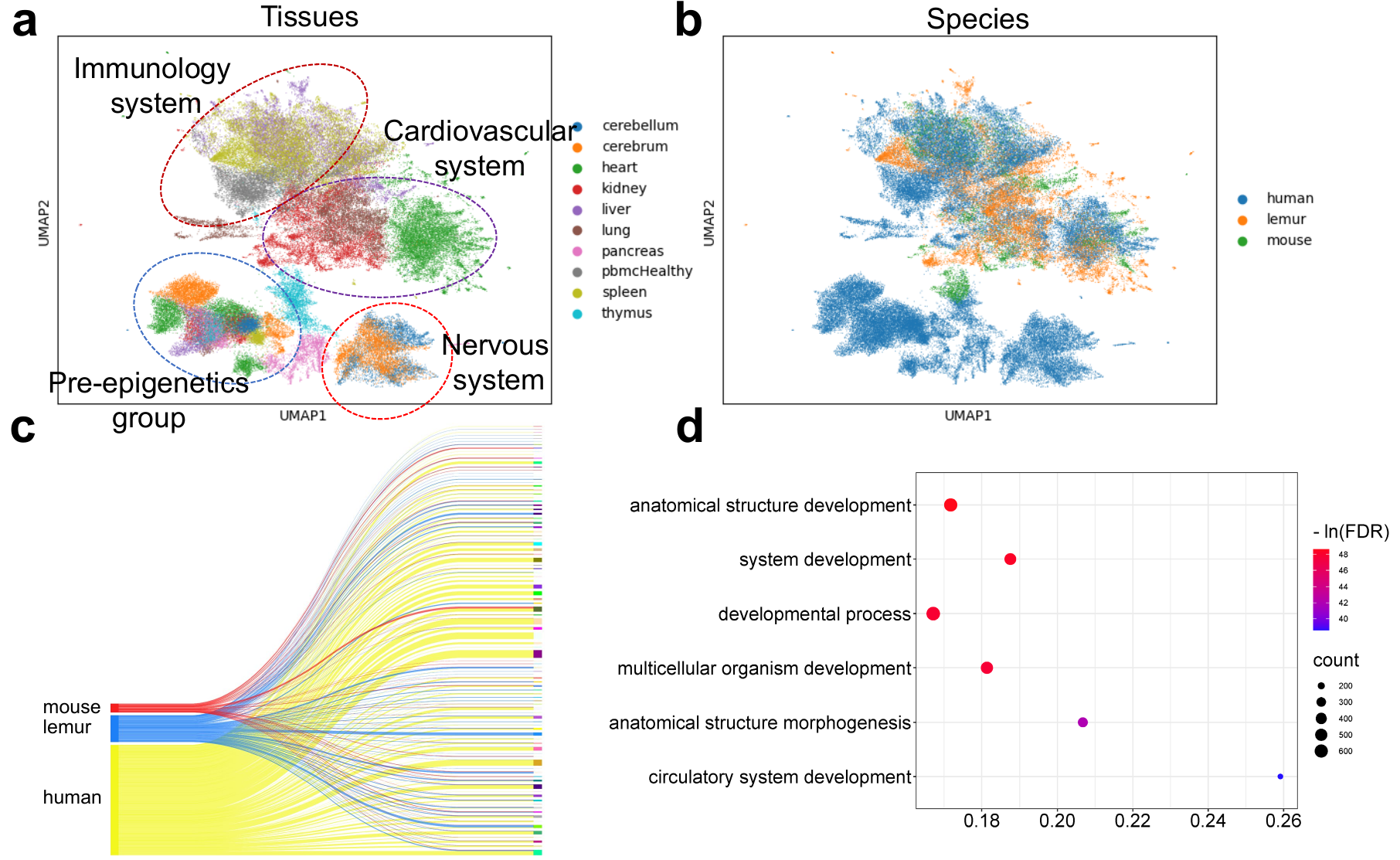}
    \caption{Gene representation learning results for multimodal biological data with different species. \textbf{(a)} represents the UMAPs of gene embeddings colored by tissue type. \textbf{(b)} represents the UMAPs of gene embeddings colored by common function groups. \textbf{(c)} is a Sankey plot \cite{sankeyplotgithub} to show the genes overlap of different species in the same clusters. \textbf{(d)} shows the top5 pathways related to the genes in the special cluster discovered by GOEA. The bubble plots in this paper were created based on ggplot2 \cite{wickham2011ggplot2}. 
    \label{fig:multi species results}}
\end{figure}
\begin{figure}[ht]
    \centering
    \includegraphics[width=0.8\textwidth]{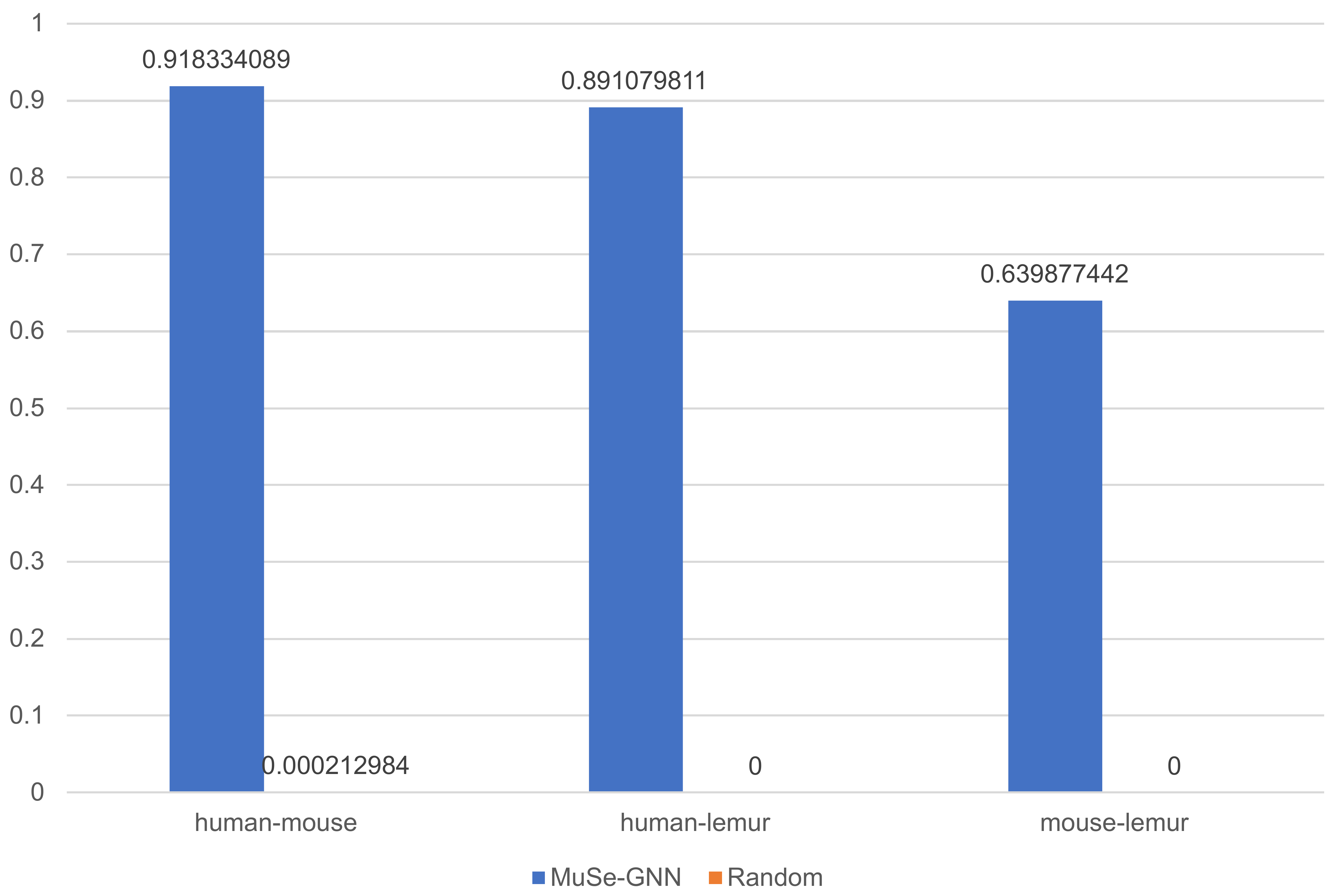}
    \caption{Compare the proportion of orthology genes discovered by MuSe-GNN and random selection. 
    \label{fig:check orhtology genes}}
\end{figure}
Integrating datasets from different species based on cells is challenging because we always need extra information, either orthology genes (genes from different species whose evaluation is only related to speciation events)  information or protein embeddings information \cite{rosen2023towards}. However, using genes as anchors simplifies this task. We examined gene embeddings generated by MuSe-GNN across three mammalian species, including humans, lemurs, and mice. As shown in Figure \ref{fig:multi species results} (a) and (b), our method effectively integrated information from different species into a shared space, preserving gene similarity across various techniques, tissues, and species. We also observed the co-embedding of genes from different species but the same tissue, supporting the similarity of evaluation direction. Since the enrichment of orthology genes can also be used as proof to evaluate whether genes with similar functions are grouped \cite{gabaldon2013functional}, we collected genes from different clusters with three species pairs and computed the proportion of these genes in the orthology genes database \cite{altenhoff2021oma}. Furthermore, we compared the results based on gene embeddings to random selection as a null approach. From Figure \ref{fig:check orhtology genes}, we could conclude that gene embeddings from our model can effectively enrich genes from different species but with similar functions.

Furthermore, Figure \ref{fig:multi species results} (c) reveals that genes from different species are distributed across multiple common function clusters. To explore shared gene functions across species, we selected one common function cluster containing genes from all three species for GOEA analysis. Since most genes in this cluster originated from heart tissue, we anticipated pathway enrichment related to multicellular systems and circulatory systems, as displayed in Figure \ref{fig:multi tissue results} (d). Thus, we demonstrated that MuSe-GNN can learn gene embeddings from multiple species and identify genes with similar functions across various species.

\section{Lung cancer analysis}
\label{appendix: lung cancer information}
In this section, we applied MuSe-GNN to analyze differentially co-expressed genes in two types of datasets, comprising lung cancer samples and healthy samples. We converted the scRNA-seq results into distinct graphs according to our graph construction strategy and trained MuSe-GNN to generate gene embeddings. We then employed the Leiden algorithm to identify common functional gene groups. After collecting essential genes with cancer-related special functions, we used GOEA and IPA to uncover more biological information.

\begin{figure}[ht]
    \centering
    \includegraphics[width=1\textwidth]{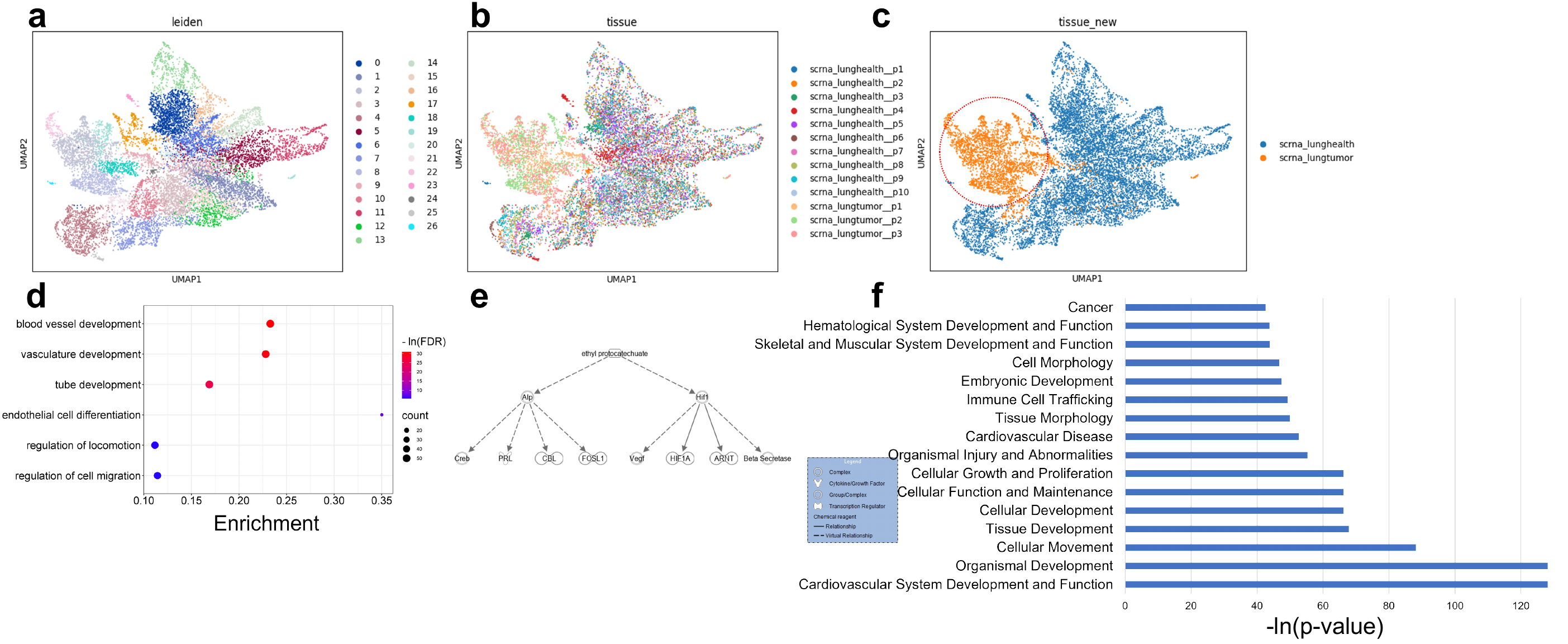}
    \caption{Gene representation learning results for samples with cancer and healthy samples. \textbf{(a)} represents the UMAPs of gene embeddings colored by functional groups. \textbf{(b)} represents the UMAPs of gene embeddings colored by datasets. \textbf{(c)} represents the gene embeddings colored by the conditions, and the red circle reflects the differential co-expression genes. \textbf{(d)} shows the top6 pathways related to the genes in the special cluster discovered by GOEA. \textbf{(e)} represents the causal network existing in the special cluster discovered by IPA. \textbf{(f)} represents the top diseases \& biological functions discovered by IPA.
    \label{fig:cancer analysis}}
\end{figure}

As shown in Figure \ref{fig:cancer analysis} (a)-(c), we observed that for lung tissue, the co-expression relationships of genes in lung tumor cells and normal cells differ significantly. Figure \ref{fig:cancer analysis} (d) displays the top pathways identified by GOEA for genes recognized as specifically co-expressed in cancer cells, with most pathways related to tissue development and cell migration. These pathways align with the fact that cancer cells often undergo multiple divisions and frequently relocate. Furthermore, using IPA, we analyzed causal networks and disease functions. Figure \ref{fig:cancer analysis} (e) illustrates an example of a causal network discovered in a specific gene group, primarily regulated by ethyl protocatechuate, indicating a chemical-dominated regulatory network. Thus, with the genes identified by MuSe-GNN, we can uncover various specific regulatory networks present in cancer and investigate strategies to inhibit cancer cell growth and metastasis. Lastly, Figure \ref{appendix: lung cancer information} (f) reveals that functions with low FDR rates include tissue development and cancer disease.

Our analysis demonstrates that MuSe-GNN can be utilized to discover functional gene clusters exhibiting specificity in cancer, assisting researchers in gaining a deeper understanding of gene expression and regulatory relationships in cancer.

\section{Gene Function Analysis}
\label{appendix: gene function analysis}
Here we continued discussing the application of gene embeddings in the gene function prediction task. We further intended to predict the characters of genes in the regulation process (as transcript factor or not) \cite{han2018trrust}. We used MuSe-GNN to generate gene embeddings for different datasets based on an unsupervised learning framework and utilized the gene embeddings as training dataset to predict the function of genes based on k-NN classifier.

\begin{figure}[ht]
    \centering
    \includegraphics[width=1\textwidth]{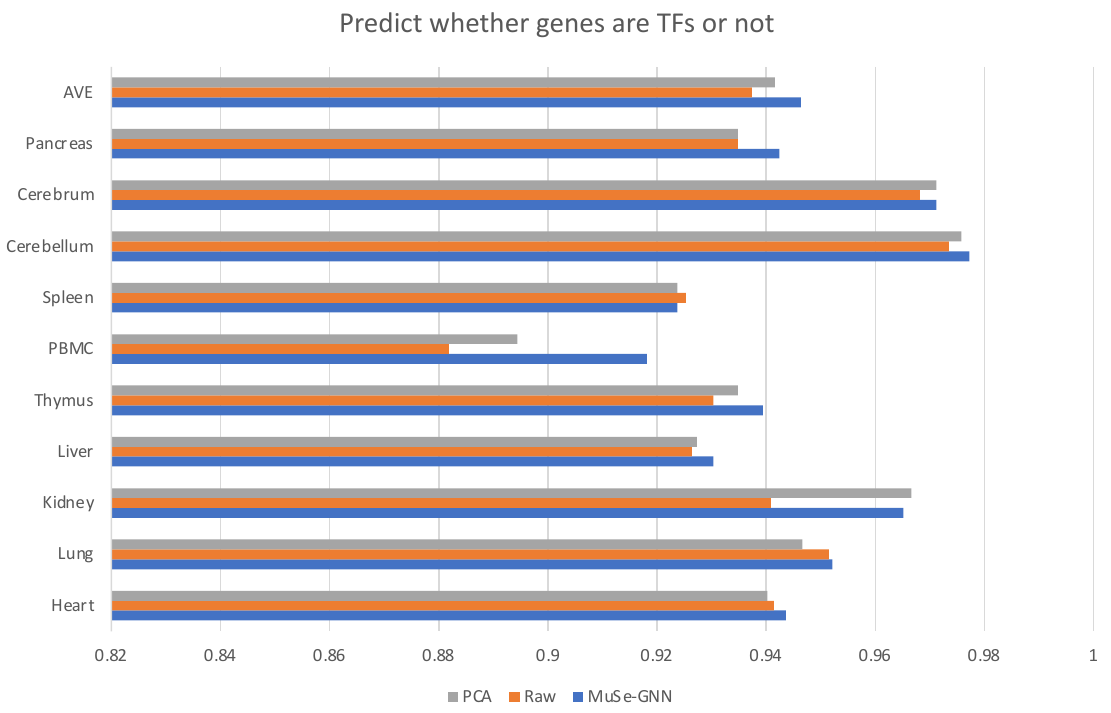}
    \caption{Accuracy for Gene-TF prediction across different tissues.
    \label{fig:gene-tf pred analysis}}
\end{figure}

In this task, we evaluated the performance of MuSe-GNN based on scRNA-seq datasets from different tissues, comparing it to the prediction results based on raw data or PCA. On average, the gene embeddings from MuSe-GNN are the most powerful embeddings, yielding highest accuracy in the validation set. Moreover, the accuracy based on gene embeddings from MuSe-GNN also surpass the score based on raw data or output of PCA in different tissues: Heart, Lung, Liver, Thymus, PBMC, Cerebellum and Pancreas. Such result is shown in Figure \ref{fig:gene-tf pred analysis}.

\section{Datasets information}
\label{appendix: ds information}
\begin{figure}[H]
    \centering
    \includegraphics[width=1\textwidth]{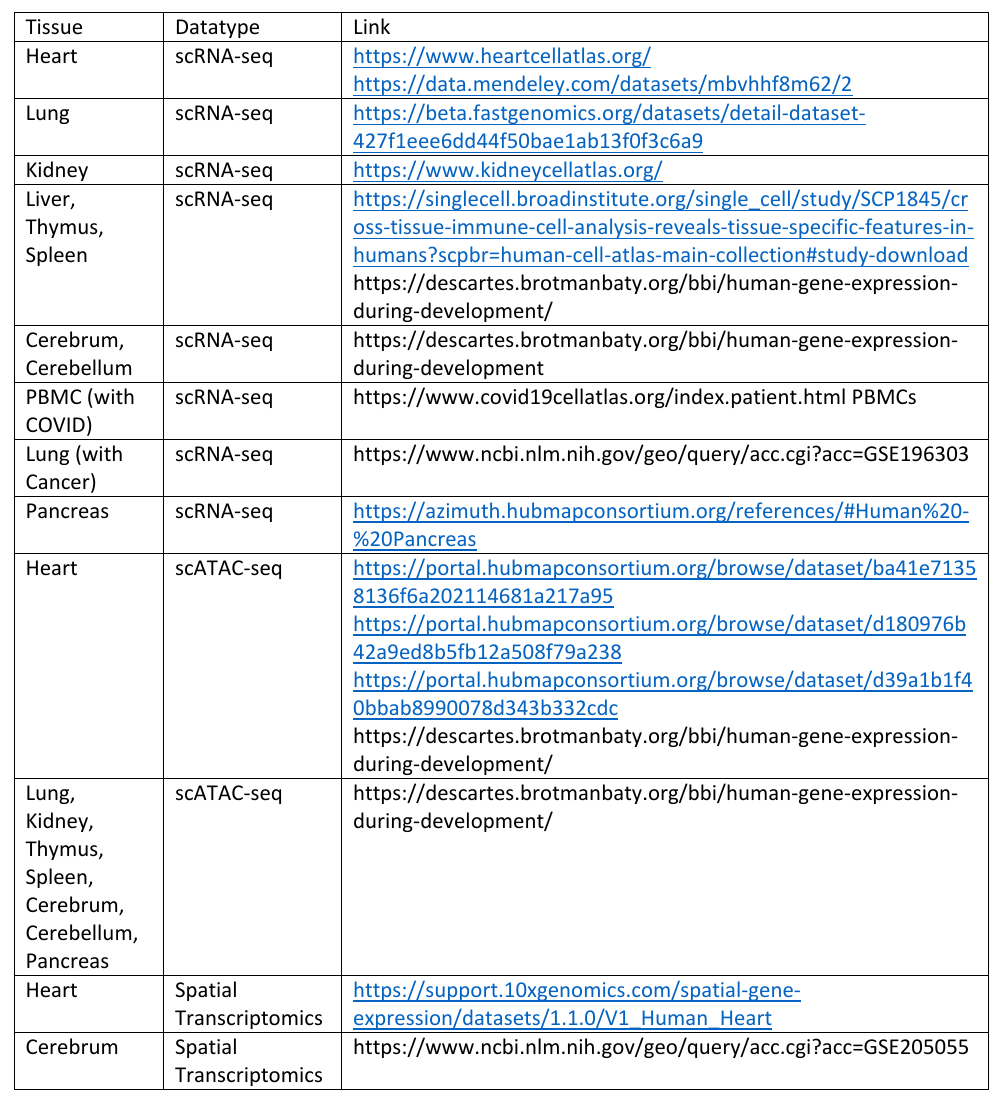}
    \caption{Information about multimodal biological datasets from Human.}
\end{figure}
\begin{figure}[H]
    \centering
    \includegraphics[width=1\textwidth]{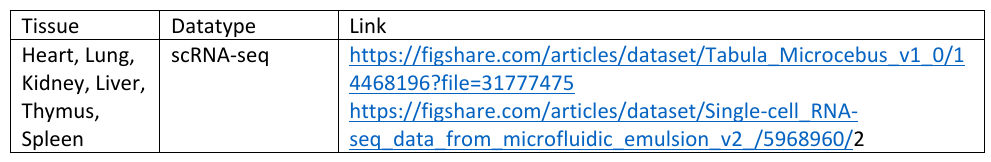}
    \caption{Information about multimodal biological datasets from Lemur and Mouse.}
\end{figure}
The statistics of different graphs can be found in Supplementary File 1.
\end{document}